\documentclass{article}

\usepackage{arxiv}

\usepackage{graphicx}
\usepackage{hyperref}

\usepackage{amssymb}
\usepackage{bm}
\usepackage{lineno}
\usepackage{amsmath}
\usepackage[ruled,linesnumbered]{algorithm2e}
\usepackage{cite}
\newcommand{\nonl}{\renewcommand{\nl}{\let\nl\oldnl}}

\SetKwComment{Comment}{$\triangleright$\ }{}
\usepackage{multirow}
\usepackage{array,color}
\usepackage{float}
\usepackage{subcaption}
\usepackage{geometry}

\title{Physics-integrated hybrid framework for model form error identification in nonlinear dynamical systems}
\author{
  Shailesh Garg  \\
  Department of Applied Mechanics\\
  Indian Institute of Technology Delhi\\
  Hauz Khas, New Delhi 110016, India. \\
  \texttt{shaileshgarg96@gmail.com} \\
  \And
  Souvik Chakraborty  \\
  Department of Applied Mechanics\\
  School of Artificial Intelligence (ScAI)\\
  Indian Institute of Technology Delhi\\
  Hauz Khas, New Delhi 110016, India. \\
  \texttt{souvik@am.iitd.ac.in} \\
 \And
  Budhaditya Hazra  \\
  Department of Civil Engineering\\
  Indian Institute of Technology Guwahati\\
  Guwahati, Assam 781039, India. \\
  \texttt{budhaditya.hazra@iitg.ac.in}
}
	
\begin{document}
\maketitle
\begin{abstract}
For real-life nonlinear systems, the exact form of nonlinearity is often not known and the known governing equations are often based on certain assumptions and approximations. Such representation introduced model-form error into the system. In this paper, we propose a novel gray-box modeling approach that not only identifies the model-form error but also utilizes it to improve the predictive capability of the known but approximate governing equation. The primary idea is to treat the unknown model-form error as a residual force and estimate it using duel Bayesian filter based joint input-state estimation algorithms. For improving the predictive capability of the underlying physics, we first use machine learning algorithm to learn a mapping between the estimated state and the input (model-form error) and then introduce it into the governing equation as an additional term. This helps in improving the predictive capability of the governing physics and allows the model to generalize to unseen environment. Although in theory, any machine learning algorithm can be used within the proposed framework, we use Gaussian process in this work. 
To test the performance of proposed framework, case studies discussing four different dynamical systems are discussed; results for which indicate that the framework is applicable to a wide variety of systems and can produce reliable estimates of original system's states. 
\end{abstract}

\section{Introduction}
Majority of the physical processes occurring in nature exhibit dynamical behaviour and thus can be modelled using linear or nonlinear differential equations.
For processes conforming to complex non-linear or chaotic behaviour, the exact form of nonlinearity is often not known and hence, one often uses parametric models to represent the same \cite{van2020modeling}. Even for cases where the exact form of nonlinearity is known, working with the same might be difficult from a computational point-of-view. Under such circumstances one often resorts to approximate methods such as linearization \cite{socha2007linearization}.
However, these approximations introduce a model form error which left unchecked can potentially result in non optimized solutions.
Naturally, it is important to develop algorithms that can estimate model form error and use it to correct the underlying approximate governing equation. 

Over the past decade or so, machine learning \cite{murphy2012machine} and deep learning \cite{goodfellow2016deep} algorithms have gained a lot of traction in various domains including aerospace \cite{kapteyn2020toward}, automobile \cite{hu2015advanced,chong2005traffic}, medical \cite{kononenko2001machine,kaur2018big}, and manufacturing \cite{wuest2016machine}. Researchers and practitioners are now looking at data-driven approaches as a possible alternative to classical physics-based approaches. Some of path-breaking work in these domains include those by \cite{qin2019data,long2018pde,long2019pde}. The basic idea in all these papers is to train data-driven deep learning algorithms to model the system. However, such approaches suffer from three major challenges. First, deep learning algorithms often need huge amount of data, which is often expensive to collect. Second, unlike physics-based models, data-driven algorithms may not generalize to unseen environment. Last but not the least, purely data-driven algorithms have limited predictive capability. For instance, if acceleration time history is provided as training data, these models are only capable of predicting acceleration time history, and it is extremely difficult to predict displacement and velocity time histories. 

For addressing issue associated with purely data-driven algorithms discussed above, physics-informed machine learning algorithms have been proposed in the literature \cite{karniadakis2021physics}. The idea here is to training the machine learning model directly from the governing equation. This is achieved by developing a physics-informed loss function. To that end, researchers have developed strong form \cite{raissi2019physics} and weak form \cite{kharazmi2021hp} based physics-informed loss functions. For time dependent systems, one can find continuous time \cite{raissi2019physics} and discrete time \cite{zhu2019physics} variants of physics-informed machine learning algorithms. Since its introduction in 2019, physics-informed machine learning algorithms have been used for solving different types of problems including fracture mechanics \cite{goswami2020transfer}, fluid mechanics \cite{sun2020surrogate}, heat transfer \cite{cai2021physics}, and reliability analysis \cite{chakraborty2020simulation}. However, these methods are only applicable when the governing equations are known in its exact form. In case the governing equation has some modeling error, the same is propagated to the physics-informed machine learning solution.

One possible alternative for eliminating model form error is the multi-fidelity framework \cite{chakraborty2021transfer,meng2020composite}. The basic premise here is first train a model based on low-fidelity data and then update it by using an auto-regressive like algorithm and few high-fidelity data. While low-fidelity data is generated from the approximate governing equation, high-fidelity data corresponds to actual field measurement or laboratory experiments. Specifically multi-fidelity schemes like co-Kriging \cite{le2014recursive,perdikaris2015multi,koziel2012variable,le2013multi} and multi-level Monte Carlo (MLMC) \cite{bierig2016approximation,giles2008multilevel,giles2017adaptive,heinrich2001multilevel} methods have been explored in the past and are well established in the available literature. Of late, nonlinear data fusion based \cite{perdikaris2017nonlinear}, transfer learning based physics-informed \cite{chakraborty2021transfer}, and Bayesian \cite{meng2021multi} multi-fidelity algorithms have been proposed. However, similar to conventional machine learning algorithms, multi-fidelity frameworks often have poor generalization and do not generalize to unseen environment.

In this paper, we propose a novel gray-box modeling framework that first estimates the model form error and then incorporate it in form of a  corrective term into the governing equation. The proposed approach utilizes Bayesian filters and machine learning algorithms to blend sensor data with known but approximate governing physics. The advantage of the proposed approach over multi-fidelity approaches are as follows:
\begin{itemize}
    \item Unlike multi-fidelity schemes, the proposed approach is capable of quantifying the model form error. As we will see later, Dual Bayesian filter is used within the proposed framework for quantifying the model form error.
    \item Unlike machine learning and multi-fidelity frameworks, the proposed approach does not attempt to replace the governing equation with a machine learning model; instead machine learning is used to enhance the known but approximate governing equation. This results in better generalization to unseen environment.
\end{itemize}
Although in theory, the proposed approach can be used with any machine learning algorithms (e.g., polynomial chaos \cite{blatman2011adaptive}, support vector machine \cite{roy2020support}, analysis-of-variance decomposition \cite{chakraborty2017polynomial}), we use Gaussian process \cite{nayek2019gaussian,bilionis2013multi} because of its proven performance in past studies conducted by the authors \cite{garg2021machine}.

The remainder of the paper is organized as follows. Section \ref{S2: DM} discusses mathematical description of the problem at hand. In Section \ref{S3: NI}, the proposed approach along with its sub-components are elaborated. We present numerical examples involving a wide class of nonlinear oscillators in Section \ref{S4: CS} to illustrate the applicability of the proposed approach. Finally, Section \ref{S5: C} provides the concluding remarks. 

\section{Problem Statement}\label{S2: DM}
\noindent Consider an $N-$DOF system subjected to deterministic and stochastic forces. The governing equation for such a system is as follows:
\begin{equation}
	\mathbf{M} \bm{\ddot X} + \mathbf{C} \bm{\dot X} + \mathbf{K} \bm X + \bm N(\bm X, \bm{\dot X}; \bm \theta_{\bm N}) = \bm F + \mathbf{\Sigma} \bm{\dot W},
\end{equation}
where $\mathbf{M}\in \mathbb R^{N\times N}$, $\mathbf{C}\in \mathbb R^{N\times N}$ and $\mathbf{K}\in \mathbb R^{N\times N}$ are the mass, damping and linear stiffness matrices respectively.
$\bm N(\bm X, \bm{\dot X}; \bm \theta_{\bm N})\in \mathbb R^{N}$ is the non-linearity vector such that $N_i$ will be the non-linear force associated with the $i-$th DOF and $\bm \theta_N$ represent the parameters controlling the behaviour of non-linearity present in the system.
$\bm X\in \mathbb R^{N}$ is the displacement vector.
$\bm F\in \mathbb R^{N}$ is the deterministic force vector and $\bm {\dot W}\in \mathbb R^{N}$ is the stochastic force vector, intensity of which is governed by the matrix $\bm \Sigma\in \mathbb R^{N\times N}$.
In a realistic scenario, the exact form of non-linearity and the exact system parameters are often not known.
Mathematically this is represented as:
\begin{equation}
	\widetilde{\mathbf{M}} \bm{\ddot \mathcal{X}} + \widetilde{\mathbf{C}} \bm{\dot \mathcal{X}} + \widetilde{\mathbf{K}} \bm {\mathcal{X}} + \widetilde{\bm N}(\bm {\mathcal{X}}, \bm{\dot \mathcal{X}}; \bm \theta_{\widetilde{\bm N}})= \bm F + \mathbf{\Sigma} \bm{\dot W},
	\label{DE-simplified model}
\end{equation}
where $\widetilde{\mathbf{M}}\in \mathbb R^{N\times N}$, $\widetilde{\mathbf{C}}\in \mathbb R^{N\times N}$ and $\widetilde{\mathbf{K}}\in \mathbb R^{N\times N}$ are the known (but not exact) parameters, and $\widetilde{\bm N}(\cdot)\in \mathbb R^N$ represent the appropriate non-linearity.
$\widetilde{\bm N}(\cdot) = 0$ represent the scenario where for the sake of simplicity non-linearity is ignored.
It should be noted that because of model form error, $\bm{\mathcal{X}}$ is only approximately equal to $\bm X$.

We now imagine a dynamical system with following governing equation ($\bm {\mathbf{M}}=\widetilde{\mathbf{M}}$ for the scope of current study):
\begin{equation}
	\mathbf{M} \bm{\ddot X} + \widetilde{\mathbf{C}} \bm{\dot X} + \widetilde{\mathbf{K}} \bm X + \widetilde{\bm N}(\bm X, \bm{\dot X}; \bm \theta_{\widetilde{\bm N}})+\bm R= \bm F + \mathbf{\Sigma} \bm{\dot W},
	\label{E:KDS:GE}
\end{equation}
where $\bm R\in \mathbb R^N$ is the residual force vector representing the model form error and can be represented as:
\begin{equation}
	\bm R = (\bm {\mathbf{C}}-\widetilde{\mathbf{C}}) \bm{\dot X} + (\bm {\mathbf{K}}-\widetilde{\mathbf{K}}) \bm X + (\bm N-\widetilde{\bm N})
	\label{E:MFE}
\end{equation}
Eq. (\ref{E:KDS:GE}) represents a dynamical system with known dynamics and unknown force vector $\bm R$.
Since, the values of $\mathbf{C}$, $\mathbf{K}$, $\bm N$ and $\bm W$ are unknown, calculating the values for $\bm R$ directly would not be possible.
The main goal of this paper is to estimate the model form error without actually knowing the exact parameters and non-linearity associated with the original system.
Also unknown are the stochastic forces acting on the system.
Accelerations or displacements or both along with the deterministic component of input forces acting on the original system are available as measurements.
The proposed algorithm should be able to adequately estimate the residual forces and map the same to the estimated states.
It should also be able to approximate the state of the system for when original system is subjected to a different loading.
\section{Proposed Framework}\label{S3: NI}
\begin{figure}[ht!]
	\centering
	\includegraphics[width = 1\textwidth]{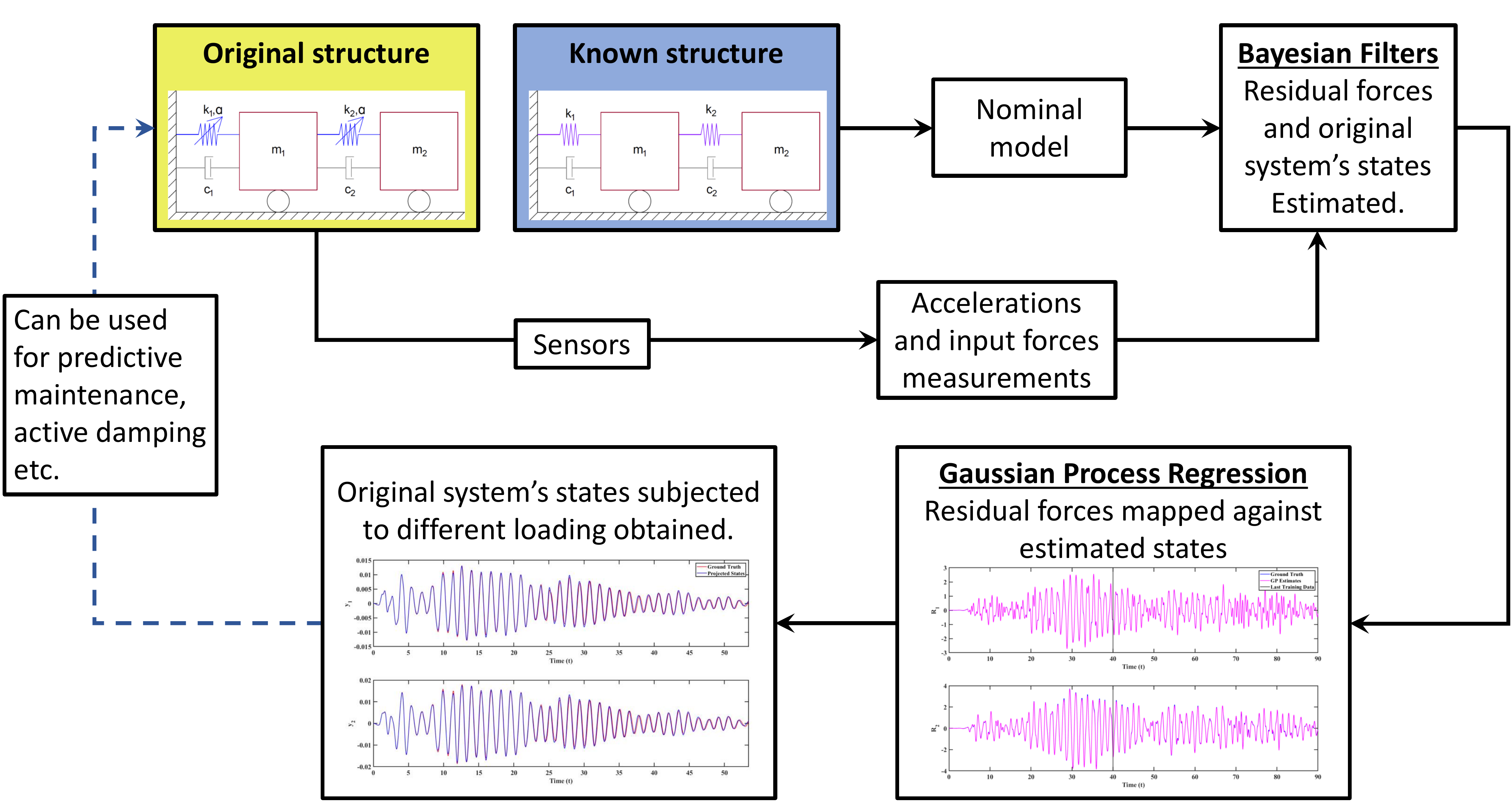}
	\caption{Schematic of the proposed framework to resolve model form uncertainty. It makes use of Bayesian filters to estimate the residual forces and then maps them to the estimated states using Gaussian process regression.}
	\label{fc}
\end{figure}

This section introduces the algorithm for the proposed framework and briefly discusses Bayesian filters and Gaussian process (GP) regression \cite{bilionis2013multi,chakraborty2021role,chakraborty2019graph} which form core of the framework.
The idea here is to jointly estimate the residual force $\bm R$ using dual Bayesian filters along with the state vector.
Thereafter, Gaussian process is used to map the estimated state vector to the estimated residual forces.
\begin{equation}\label{eq:map}
    \bm R = f(\bm{X},\bm{\dot X})
\end{equation}
Eq. (\ref{E:KDS:GE}) can then be modified as,
\begin{equation}
  {\mathbf{M}} \bm{\ddot X} + \widetilde{\mathbf{C}} \bm{\dot X} + \widetilde{\mathbf{K}} \bm X + \widetilde{\bm N} + f(\bm X,\bm {\dot X}) = \bm F+\Sigma \bm W
  \label{DE-final}
\end{equation}
Since, values for $\widetilde{\mathbf{C}}$, $\widetilde{\mathbf{K}}$, $\widetilde{\bm N}$ and $\bm F$ in Eq. (\ref{DE-final}) are now known, this becomes a \textit{forward problem} which can be solved to obtain the system states corresponding to the original system. With such a setup, the proposed approach is able to generalize to unknown environment (unseen forces).
A schematic of the proposed framework has been shown in Fig. \ref{fc}. A high level algorithm of the proposed approach is provided in Algorithm \ref{alg-1}.
\begin{algorithm}[ht!]
	\caption{High-level algorithm for the proposed framework}\label{alg-1}
	\textbf{Input:} Governing equation (approximate) and parameters (approximate) of the system under consideration. \\
	Estimate the residual force $\bm R$ and the state vectors using DBF using joint input-state estimation algorithm.\\
	Map the estimated states to the estimated residual force.\Comment*[r]{Eq. \eqref{eq:map}}
	Update the known governing equation (approximate) by including the machine learning model.\Comment*[r]{Eq. \eqref{DE-final}}
	\textbf{Outcome: }A gray-box model capable of predicting responses corresponding to different operating conditions.
\end{algorithm}

The algorithm proposed above has two key components: (a) an algorithm for jointly estimating the input and state and (b) a machine learning algorithm for mapping the estimated states and inputs. In this work, we propose to use DBF for jointly estimating the input and state vectors. As for learning the mapping between the estimated state and the force vectors, we use Gaussian process \cite{chakraborty2019graph,nayek2019gaussian} because of its already proven performance.

Details on how Bayesian filter and Gaussian process are used within the proposed framework is discussed next.
\subsection{Bayesian filters}
\begin{figure}[ht!]
    \centering
    \includegraphics[width=0.9\textwidth]{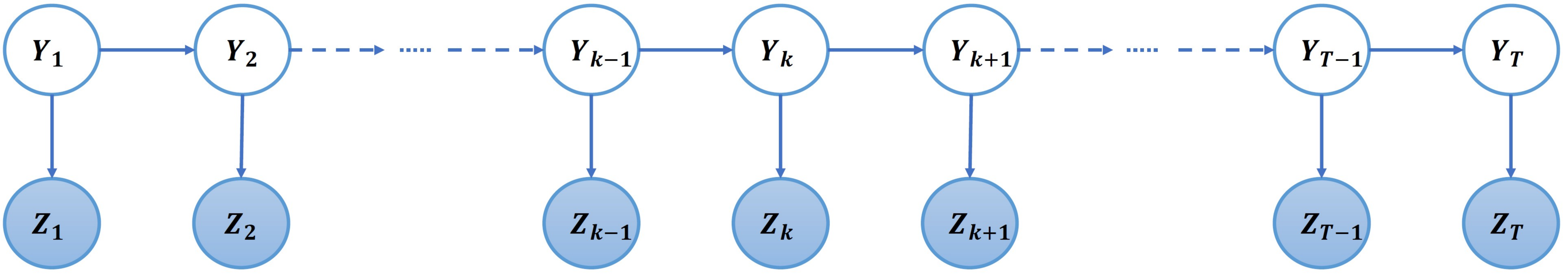}
    \caption{Probabilistic graphical model for state space model. Because of  Markovian assumption, the hidden variable $\bm Y_t$ is only dependent on $\bm Y_{t-1}$ and the observation $\bm Z_t$ is dependent on current state only.}
    \label{fig:pgm}
\end{figure}

\begin{figure}[ht!]
	\begin{subfigure}{1\textwidth}
		\centering
		\includegraphics[width = 1\textwidth]{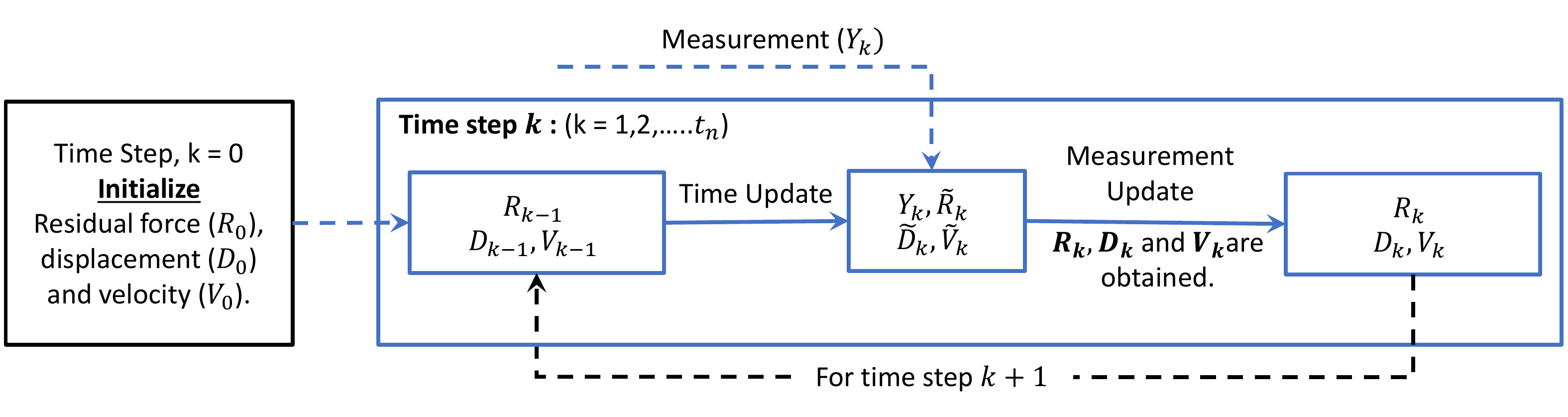}
		\caption{Process flow: Bayesian filter}
	\end{subfigure}
	\begin{subfigure}{1\textwidth}
		\centering
		\includegraphics[width = 1\textwidth]{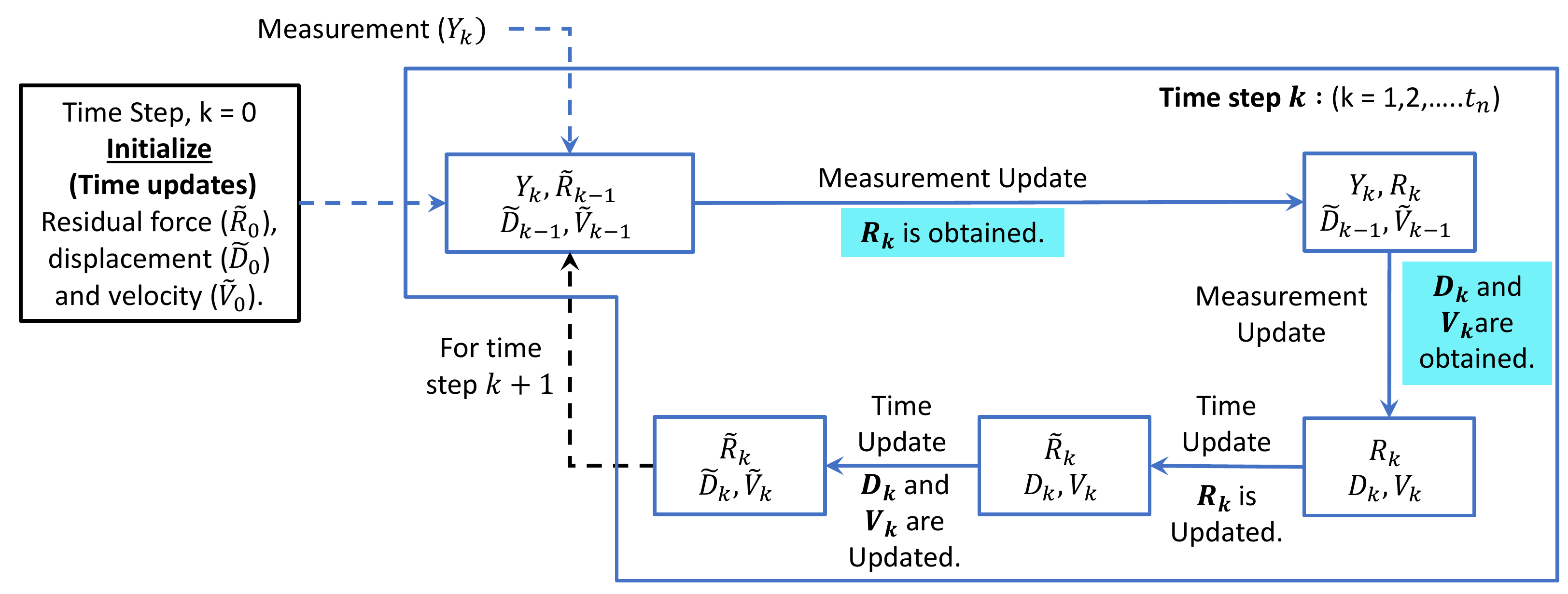}
		\caption{Process flow: Dual Bayesian filter}
	\end{subfigure}
	\caption{Comparison between process flow of Bayesian and Dual Bayesian filter.}
	\label{f-df-comp}
\end{figure}

Bayesian filters (BFs) work on the principals of Bayesian statistics and aim at estimating the hidden states given some observations.
BFs are used in a variety of domains ranging from target tracking, GPS to health industry.
The recursive nature of Bayesian filter which makes them useful for large data sets, is possible because of the Markovian assumption wherein the current state of system is assumed to be influenced by previous state only and the current measurement is assumed to be affected by current state only. Rest of the state histories can therefore be ignored while analysis, reducing the computational cost drastically.
A probabilistic graphical model representing the first-order Markov assumption is shown in Fig. \ref{fig:pgm}.
Kalman filter \cite{sarkka2013bayesian,welch1995introduction,chen2003bayesian}, Extended Kalman filter \cite{sarkka2013bayesian}, Unscented Kalman filter \cite{wan2000unscented} are special types of recursive Bayesian filters, each catering to different types of filter model.
KF takes on linear filter models while EKF and UKF are used for non-linear filter model.
UKF differs from EKF in the way that it uses unscented transform instead of linearization to analyze the non-linear models.
While KF, EKF and UKF individually can be used for force state estimation, dual Bayesian filters give better convergence and thus produce better estimates.
Differences between the process flow of Bayesian and dual Bayesian filters are shown using schematic in Fig. \ref{f-df-comp}.
While conventional filters augment unknown states and forces in a single vector of the form $[X,\,\dot X,\,R]^T$, dual filters makes two separate state-space models.
One of the models outlines the behaviour of force while the other is based on system states.
A generic state space model for DBF is as follows:
\begin{equation}
    \begin{array}{c}
    \bm{\mathrm f}_k\sim p(\bm{\mathrm f}_k|\bm{\mathrm f}_{k-1})\\
    \bm{\mathrm y}_k\sim p(\bm{\mathrm y}_k|\bm{\mathrm y}_{k-1},\bm{\mathrm f}_{k-1})\\
    \bm{\mathrm z}_k\sim p(\bm{\mathrm z}_k|\bm{\mathrm y}_{k},\bm{\mathrm f}_{k}),
    \end{array}
    \label{eq-dbf}
\end{equation}
where $\bm{\mathrm f}_k$ is the unknown force vector at time step $k$, $\bm{\mathrm y}$ is the state vector representing state of system at time step $k$.
$\bm{\mathrm z}_k$ is the measurement vector observed at time step $k$.
$p(\bm{\mathrm f}_k|\bm{\mathrm f}_{k-1})$ is the force model describing the trend of the unknown forces.
$\bm{\mathrm y}_k\sim p(\bm{\mathrm y}_k|\bm{\mathrm y}_{k-1},\bm{\mathrm f}_{k-1})$ is the dynamic model describing the dynamics of the system and $p(\bm{\mathrm z}_k|\bm{\mathrm y}_{k},\bm{\mathrm f}_{k})$ is the measurement model, which shows the distribution of measurements given the state of system.

\noindent \textbf{Remark 1: }We note that DBF is a generic term and can encompass a wide array of algorithms. The choice of DBF algorithm depends on the system at hand. In this paper, we propose to use Dual Kalman Filter (DKF) \cite{dertimanis2019input} when the known governing equation (approximate) is linear. On the other hand, if the known governing equation is nonlinear, we use Dual Unscented Kalman Filter (DUKF) \cite{gove2006application}.

DKF is a popular algorithm that is used in literature for joint input-state estimation. The filtering model in DKF is expressed as follows:
\begin{equation}
  \begin{array}{c}
  \bm{\mathrm{f}}_k = \mathbf{T}\bm{\mathrm{f}}_{k-1}+\bm q^1_{k-1}\\
  \bm{\mathrm{y}}_k = \mathbf{A}_s\bm{\mathrm{y}}_{k-1}+\mathbf{A}_i\bm F^i_{k-1}+\mathbf{A}_f\bm{\mathrm{f}}_{k-1}+\bm q^2_{k-1}\\
  \bm{\mathrm{z}}_k = \mathbf{H}_s\bm{\mathrm{y}}_{k}+\mathbf{H}_f\bm{\mathrm{f}}_{k}+\bm r_{k}
  \end{array}
  \label{eq-dkf}
\end{equation}
where the state vector is defined as $\bm{\mathrm{y}} = [\bm X,\,\bm {\dot X}]^T$. $\bm q^1\sim N(0,\mathbf{Q}^1)$ and $\bm q^2\sim N(0,\mathbf{Q}^2)$ are the process noises with co-variance $\mathbf{Q}^1$ and $\mathbf{Q}^2$ respectively. $\bm r\sim N(0,\mathbf{R})$ is the measurement noise with noise co-variance $\mathbf{R}$. $\bm F^i$ are the known input forces acting on the system. Process flow for DKF is given in Algorithm \ref{alg-2}.
Matrices $\mathbf A_i$ relate the previous state dynamics to the current state whereas $\mathbf H_i$ relate the current state of the system to the current observed measurement.
Specifically $\mathbf A_s$ maps the previous state to the current state and $\mathbf A_f,\,\mathbf A_i$ map the unknown force and the input forces to the current state.
Similarly $\mathbf H_s$ and $\mathbf H_f$  map the current state and force to the current measurement. 
\begin{algorithm}[ht!]
	\caption{DKF Algorithm}\label{alg-2}
	\textbf{Initialize}: Predicted state $\bm p_s$, covariance of  predicted state $\mathbf{C}_{p_s}$, predicted force $\bm p_f$, and covariance of predicted force $\mathbf{C}_{p_f}$\\
	Repeat steps 3-14 for time step $k=1,2,....,$end\\
	\tcp{Measurement Update - I (3-6)}
	$\bm e_f = \bm{\mathrm{z}}[k]-(\mathbf{H}_s\bm p_s+\mathbf{H}_f\bm p_f)$\\
	$\mathbf{K}_f=(\mathbf{H}_f\mathbf{C}_{p_f}\mathbf{H}_f^T+R)^{-1}\mathbf{H}_f\mathbf{C}_{p_f}$\\
	$\bm{\mathrm{f}}[k]=\bm c_f=\bm p_f+\mathbf{K}_f\bm e_f$ \Comment*[r]{Force estimated at time step k}
	$\mathbf{C}_{c_f}=\mathbf{C}_{p_f}-\mathbf{K}_f\mathbf{H}_f\mathbf{C}_{p_f}$\\
	\tcp{Measurement Update - II (7-10)}
	$\bm e_s = \bm{\mathrm{z}}[k]-(\mathbf{H}_s\bm p_s+\mathbf{H}_f\bm c_f)$\\
	$\mathbf{K}_s=(\mathbf{H}_s\mathbf{C}_{p_s}\mathbf{H}_s^T+\mathbf{R})^{-1}\mathbf{H}_s\mathbf{C}_{p_s}$\\
	$\bm{\mathrm{y}}[k]=\bm c_s=\bm p_s+\mathbf{K}_s\bm e_s$ \Comment*[r]{States estimated at time step k}
	$\mathbf{C}_{c_s}=\mathbf{C}_{p_s}-\mathbf{K}_s\mathbf{H}_s\mathbf{C}_{p_s}$\\
	\tcp{Time Update - I (11-12)}
	$\bm p_f=\mathbf{T}\bm c_f$\\
	$\mathbf{C}_{p_f}=\mathbf{T}\mathbf{C}_{c_f}\mathbf{T}^T+\mathbf{Q}^1$\\
	\tcp{Time Update - II (13-14)}
	$\bm p_s=\mathbf{A}_s\bm c_s+\mathbf{A}_i\bm F^i[k]+\mathbf{A}_f\bm c_f$\\
	$\mathbf{C}_{p_s}=\mathbf{A}_s\mathbf{C}_{c_s}\mathbf{A}_s^T+\mathbf{Q}^2$. \\
	\textbf{Outcome:} Estimated model-form error and state vectors.
\end{algorithm}

The nonlinear counterpart of DKF is the Dual Unscented Kalman Filter (DUKF). The filter equations in DUKF can be written as:
\begin{equation}
  \begin{array}{c}
  \bm{\mathrm{f}}_k = f^1(\bm{\mathrm{f}}_{k-1})+\bm q^1_{k-1}\\
  \bm{\mathrm{y}}_k = f^2(\bm{\mathrm{y}}_{k-1},\bm F^i_{k-1},\bm{\mathrm{f}}_{k-1})+\bm q^2_{k-1}\\
  \bm{\mathrm{z}}_k = h(\bm{\mathrm{y}}_{k},\bm{\mathrm{f}}_{k})+\bm r_{k},
  \end{array}
  \label{eq-dukf}
\end{equation}
where $f^i(\cdot)$ and $h(\cdot)$ are the dynamic and measurement model respectively.
We approximate the filtering distributions of unknown force and state dynamic models described in Eq. \eqref{eq-dukf} as:
\begin{equation}
\begin{array}{c}
p(\bm{\mathrm{f}}_k|\bm{\mathrm{z}}_{1:k})\simeq N(\bm{\mathrm{f}}_k|\bm m_{1_k},\mathbf P_{1_k})\\
p(\bm{\mathrm{y}}_k|\bm{\mathrm{z}}_{1:k})\simeq N(\bm{\mathrm{y}}_k,\bm{\mathrm{f}}_k|\bm m_{2_k},\mathbf P_{2_k})
\end{array}
\end{equation}
where $\bm m_i$ and $\mathbf P_i$ are the mean and co-variance matrices governing the properties of the distribution.
DUKF used in this paper follows the same trend as that followed by DKF i.e., measurement updates for unknown force and states will be followed by their respective time updates. However, because of the presence of UKF within the DUKF framework, the computation is more involved. Overall, the computation carried out inside the DUKF algorithm can be divided into two steps (a) computation of the DUKF weights and (b) joint estimation of the input and state vectors using DUKF. We present calculation of DUKF weights in Algorithm \ref{alg-dukf-wc} and the overall DUKF algorithm in Algorithm \ref{alg-3}.

\begin{algorithm}[ht!]
	\caption{DUKF weights calculation}\label{alg-dukf-wc}
	\tcp{For unknown force vector}
	\textbf{Input:} Length of unknown force vector $L_f$ and Length of state vector $L_s$. \\
	$\alpha_1 \leftarrow 1$, $\alpha_2 \leftarrow 1$, $\beta_1 \leftarrow 2$, $\beta_2 \leftarrow 2$, $\kappa_1 \leftarrow 0$, $\kappa_2 \leftarrow 0$\\
	$\lambda_1 = \alpha_1^2(L_f+\kappa_1)-L_f$\\
	$W_{1_m}^i = \displaystyle\frac{\lambda_1}{L_f+\lambda_1}$\Comment*[r]{for $i=0$}
	$W_{1_c}^i = \displaystyle\frac{\lambda_1}{L_f+\lambda_1}+(1-\alpha_1^2+\beta_1)$\Comment*[r]{for $i=0$}
	$W_{1_m}^i = \displaystyle\frac{1}{2(L_f+\lambda_1)}$\Comment*[r]{for $i=1,....,2L_f$}
	$W_{1_c}^i = W_{1_m}^i$\Comment*[r]{for $i=1,....,2L_f$}
    \tcp{For unknown state vector}
	$\lambda_2 = \alpha_2^2(L_s+\kappa_2)-L_s$\\
	$W_{2_m}^i = \displaystyle\frac{\lambda_2}{L_s+\lambda_2}$\Comment*[r]{for $i=0$}
	$W_{2_c}^i = \displaystyle\frac{\lambda_2}{L_s+\lambda_2}+(1-\alpha_2^2+\beta_2)$\Comment*[r]{for $i=0$}
	$W_{2_m}^i = \displaystyle\frac{1}{2(L_s+\lambda_2)}$\Comment*[r]{for $i=1,....,2L_s$}
	$W_{2_c}^i = W_{2_m}^i$\Comment*[r]{for $i=1,....,2L_s$}
	\textbf{Output:} Calculated weights for sigma points.
\end{algorithm}

\begin{algorithm}[ht!]
    \small
	\caption{\small DUKF algorithm}\label{alg-3}
	Calculate DUKF weight.\Comment*[r]{Alg. \ref{alg-dukf-wc}}
	 \textbf{Initialize:} Predicted mean $\bm m_{1_0}^-$, $\bm m_{2_0}^-$ and co-variance $\mathbf P _{1_0}^-$, $\mathbf P _{2_0}^-$ \\
	$\bm m_{1_k}^- = \bm m_{1_0}^-,\,\bm m_{2_k}^- = \bm m_{2_0}^-,\,\mathbf P_{1_k}^- = \mathbf P_{1_0}^-,\,\mathbf P_{2_k}^- = \mathbf P_{2_0}^-$\\
	{\nonl for $k = 1,2,....,t_n$}\\
	$\mathbf{\mathcal{F}}_k^- = [\bm m_{1_{k-1}}^-\,\,\,\,\,\bm m_{1_{k-1}}^-+\sqrt{L_f+\lambda_1}\left[\sqrt{\mathbf P_{1_{k-1}}^-}\right]\,\,\,\,\,\bm m_{1_{k-1}}^--\sqrt{L_f+\lambda_1}\left[\sqrt{\mathbf P_{1_{k-1}}^-}\right]]$\\
	$\bm{\mathcal{Z}}_{1_k}^i=h(\bm m_{2_{k-1}}^-,\bm{\mathcal{F}}_k^{-i})$\Comment*[r]{for $i = 0,1,....,2L_f$}
	$\bm{\mu}_{1_k}=\displaystyle\sum_{i=0}^{2L_f}W_{1_m}^i\bm{\mathcal{Z}}_{1_k}^i$\\
	$\mathbf S_{1_k}=\displaystyle\sum_{i=0}^{2L_f}W_{1_c}^i(\bm{\mathcal{Z}}_{1_k}^i-\bm \mu_{1_k})(\bm{\mathcal{Z}}_{1_k}^i-\bm \mu_{1_k})^T+\mathbf R_k$\\
	$\mathbf C_{1_k}=\displaystyle\sum_{i=0}^{2L_f}W_{1_c}^i(\bm{\mathcal{F}}_k^{-i}-\bm m_{1_{k-1}}^-)(\bm{\mathcal{Z}}_{1_k}^i-\bm \mu_{1_k})^T$\\
	$\mathbf K_{1_k}=\mathbf C_{1_k}\mathbf S_{1_k}^{-1}$\\
	$\bm m_{1_k} = \bm m_{1_{k-1}}^-+\mathbf K_{1_k}(\bm{\mathrm{z}}_k-\bm \mu_{1_k})$;\,\,\,\,\,$\mathbf P_{1_k} = \mathbf P_{1_{k-1}}^--\mathbf K_{1_k}\mathbf S_{1_k}\mathbf K_{1_k}^T$\Comment*[r]{Estimated force mean and co-variance}
	$\mathbf{\mathcal{Y}}_k^- = [\bm m_{2_{k-1}}^-\,\,\,\,\,\bm m_{2_{k-1}}^-+\sqrt{L_s+\lambda_2}\left[\sqrt{\mathbf P_{2_k}^-}\right]\,\,\,\,\,\bm m_{2_{k-1}}^--\sqrt{L_s+\lambda_2}\left[\sqrt{\mathbf P_{2_k}^-}\right]]$\\
	$\bm{\mathcal{Z}}_{2_k}^i=h(\bm{\mathcal{Z}}_k^{-i},\bm m_{1_k})$\Comment*[r]{for $i = 0,1,....,2L_s$}
	$\bm \mu_{2_k}=\displaystyle\sum_{i=0}^{2L_s}W_{2_m}^i\bm{\mathcal{Z}}_{2_k}^i$\\
	$\mathbf S_{2_k}=\displaystyle\sum_{i=0}^{2L_s}W_{2_c}^i(\bm{\mathcal{Z}}_{2_k}^i-\bm \mu_{2_k})(\bm{\mathcal{Z}}_{2_k}^i-\bm \mu_{2_k})^T+\mathbf R_k$\\
	$\mathbf C_{2_k}=\displaystyle\sum_{i=0}^{2L_s}W_{2_c}^i(\bm{\mathcal{Y}}_k^{-i}-\bm m_{2_{k-1}}^-)(\bm{\mathcal{Z}}_{2_k}^i-\bm \mu_{2_k})^T$\\
	$\mathbf K_{2_k}=\mathbf C_{2_k}\mathbf S_{2_k}^{-1}$\\
	$\bm m_{2_k} = \bm m_{2_{k-1}}^-+\bm K_{2_k}(\bm{\mathrm{z}}_k-\bm \mu_{2_k})$;\,\,\,\,\,$\mathbf P_{2_k} = \mathbf P_{2_{k-1}}^--\mathbf K_{2_k}\mathbf S_{2_k}\mathbf K_{2_k}^T$ \Comment*[r]{Estimated state mean and co-variance}
	$\mathbf{\mathcal{F}}_k = [\bm m_{1_k}\,\,\,\,\,\bm m_{1_k}+\sqrt{L_f+\lambda_1}\left[\sqrt{\mathbf P_{1_k}}\right]\,\,\,\,\,\bm m_{1_k}-\sqrt{L_f+\lambda_1}\left[\sqrt{\mathbf P_{1_k}}\right]]$\\
	$\bm{\mathbb{F}}_k^i=f^1(\bm{\mathcal{F}}_k^{-i})$\Comment*[r]{for $i = 0,1,....,2L_f$}
	$\bm m_{1_k}^- =\displaystyle\sum_{i=0}^{2L_f}W_{1_m}^i\bm{\mathbb{F}}_k^i$;\,\,\,\,\,$\mathbf P_{1_k}^-=\displaystyle\sum_{i=0}^{2L_f}W_{1_c}^{(i)}(\bm{\mathbb{F}}_k^i-\bm m_{1_k}^-)(\bm{\mathbb{F}}_k^i-\bm m_{1_k}^-)^T+\mathbf Q^1_k$\\
	$\mathbf{\mathcal{Y}}_k = [\bm m_{2_k}\,\,\,\,\,\bm m_{2_k}+\sqrt{L_f+\lambda_2}\left[\sqrt{\mathbf P_{2_k}}\right]\,\,\,\,\,\bm m_{2_k}-\sqrt{L_f+\lambda_2}\left[\sqrt{\mathbf P_{2_k}}\right]]$\\
	$\bm{\mathbb{Y}}_k^i=f^2(\bm{\mathcal{Y}}_k^{-i})$\Comment*[r]{for $i = 0,1,....,2L_s$}
	$\bm m_{2_k}^- =\displaystyle\sum_{i=0}^{2L_s}W_{2_m}^i\bm{\mathbb{Y}}_k^i$;\,\,\,\,\,$\mathbf P_{2_k}^-=\displaystyle\sum_{i=0}^{2L_s}W_{2_c}^{(i)}(\bm{\mathbb{Y}}_k^i-\bm m_{2_k}^-)(\bm{\mathbb{Y}}_k^i-\bm m_{2_k}^-)^T+\mathbf Q^2_k$\\ 
	\textbf{Output:} Estimated model-form error and state vectors
\end{algorithm}

For using either DKF or DUKF in practice, we need to estimate the filter models for a given system. Details on filter equations are formulated from governing equation is well-documented in the literature \cite{sarkka2013bayesian}. Nonetheless, for details on the same, interested readers may refer \ref{app:1} (linear system) and \ref{app:2} (nonlinear system).
\subsection{Gaussian process regression}\label{GPR}
Gaussian process regression \cite{nayek2019gaussian,bilionis2013multi} is a popular non-parametric machine learning based regression technique. Being data driven, GPR does not require a prior physical model and hence can be used for a wide range of problems.
Advantage of GPR over conventional regression techniques is that along with estimating the data it gives the uncertainty attached to it which can help the user make better decisions while using the data for any particular application.
In this paper GPR has been used to map the estimated states $\bm{\mathrm{y}}$ to the residual forces $R$ and obtain a GPR model.
The states and forces used to prepare the GPR models are obtained from DBFs as discussed earlier.

\begin{figure}[ht!]
    \centering
    \includegraphics[width = 1\textwidth]{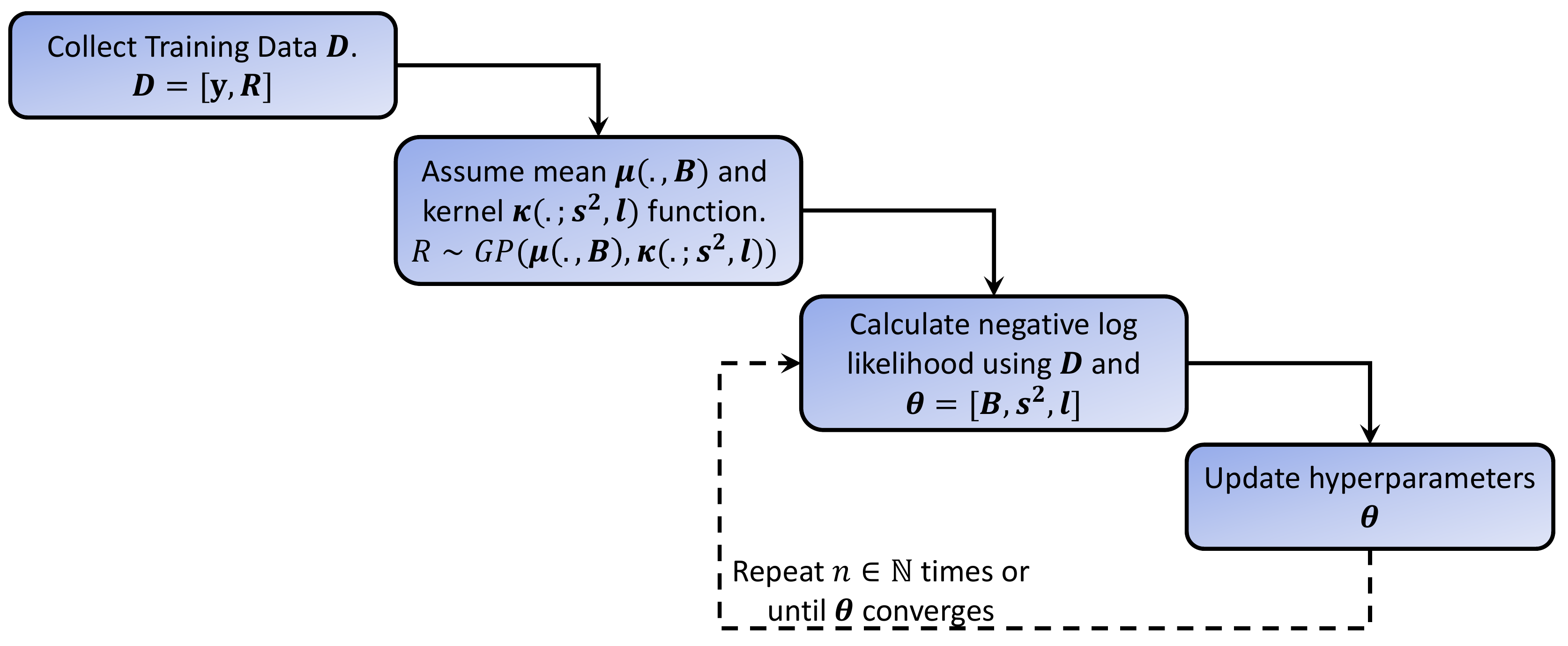}
    \caption{Schematic for GPR training process.}
    \label{f:GPR}
\end{figure}

GPR treats the whole data as a Gaussian process with some mean and co-variance.
As such before starting the regression process a mean and co-variance function are assumed and the data is idealized as follows:
\begin{equation}
  \bm R\sim \mathcal{GP}(\bm\mu(\bm{\mathrm{y}};B),\bm\kappa(\bm{\mathrm{y}},\bm{\mathrm{y}}';s^2\mathbf{I},l),
\end{equation}
where the state vector $\bm{\mathrm{y}}$ acts like a independent input variable such that $\bm R=g(\bm{\mathrm{y}})$.
$\mu(.,B)$ is the mean function which can be assumed as a constant or some function of the input variable.
$\kappa(.,.,s^2\mathbf{I})$ is the co-variance or kernel function which is assumed based on the prior knowledge of the data being analyzed.
A lot of predefined kernel functions exist in the literature namely squared exponential, Matern 32, Matern 52, etc.
Upon selection of mean and co-variance functions, training data $\mathcal D = [\bm{\mathrm{y}},\,\bm R]$ is used to optimize the hyper-parameters $B$, $s$, and $l$.
This is done by maximizing the likelihood of the training data.
A schematic for GPR training process is shown in Fig. \ref{f:GPR}.

For predicting response at a new point $\bm{\mathrm{y}}^*$, we utilize the fact that the joint distribution between data at known input points $\bm{\mathrm{y}}$ and unknown input points $\bm{\mathrm{y}}^*$ under GPR assumption is as follows:
\begin{equation}
    \left[\begin{matrix}\bm R\\ \bm R^*\end{matrix}\right]\sim\mathcal{N}\left(\left[\begin{matrix}\bm\mu(\bm{\mathrm{y}})\\ \bm\mu(\bm{\mathrm{y}}^*)\end{matrix}\right],\left[\begin{matrix}\bm\kappa(\bm{\mathrm{y}},\bm{\mathrm{y}}) & \bm\kappa(\bm{\mathrm{y}},\bm{\mathrm{y}}^*)\\ \bm\kappa(\bm{\mathrm{y}}^*,\bm{\mathrm{y}}) & \bm\kappa(\bm{\mathrm{y}}^*,\bm{\mathrm{y}}^*)\end{matrix}\right]\right)
\end{equation}
Using properties of Gaussian process, it can then be shown that
$\bm R^*$ is also a Gaussian process with mean $E[\bm R^*|\bm R]$ and co-variance $Cov[\bm R^*|\bm R]$,
\begin{equation}
  \begin{array}{c}
  E[\bm R^*|\bm R] = \bm\mu(\bm{\mathrm{y}}^*)+\kappa(\bm{\mathrm{y}}^*,\bm{\mathrm{y}})\,\kappa(\bm{\mathrm{y}},\bm{\mathrm{y}})^{-1}\,(\bm R-\bm\mu(\bm{\mathrm{y}})) \\
  Cov[\bm R^*|\bm R] = \kappa(\bm{\mathrm{y}}^*,\bm{\mathrm{y}}^*)-\kappa(\bm{\mathrm{y}}^*,\bm{\mathrm{y}})\,\kappa(\bm{\mathrm{y}},\bm{\mathrm{y}})^{-1}\,\kappa(\bm{\mathrm{y}}^*,\bm{\mathrm{y}})^T 
  \end{array}
  \label{eq-gpr-mean-cov}
\end{equation}
The mean and co-variance formulae in Eq. (\ref{eq-gpr-mean-cov}) can be easily modified if there is noise in the given training data \cite{sarkka2019applied}.
\section{Numerical Examples}\label{S4: CS}\noindent
In this section, we present three numerical examples to illustrate the performance of the proposed approach. The examples considered involve popular nonlinear oscillators such as duffing oscillator, Bouc-Wen oscillator and duffing Van-der Pol oscillator. For the first two examples, we have assumed the known system to be linear and hence, DKF in conjunction with GP has been used. For the third example, we show the performance of the proposed approach when the known system is also nonlinear. Case I and II take accelerations and input forces as measurements while case-III takes both acceleration and displacement along with input forces as measurement. As discussed earlier, the proposed algorithm uses the measurements to model the model-form error as a residual force and produces a GPR model for the same; this is referred to as `residual force model' or `RF model' in the following text. To illustrate the robustness of the proposed approach, we examine the predictive capability when the underlying system is subjected to a completely different forcing function, referred to here as `different input'.




We also examine the performance of the proposed approach outside the training window.
\subsection{Case-I : 2-DOF system with duffing oscillators}
\begin{table}[ht!]
\caption{System parameters for Case-I.}
	\centering\scalebox{0.85}{
		\begin{tabular}{|c|c|c|c|c|}
			\hline
			System & Mass (Kg) & Stiffness (N/m) & Damping (Ns/m) & Non-linear Parameters \\\hline 
			Original & $m_1 = 30,m_2 = 15$ & $k_1 = 1000,k_2 = 1000$ & $c_1 = 10,c_2 = 5$ & $\alpha_{do} = 100$ \\
			Known & $m_1 = 30,m_2 = 15$ & $\tilde k_1 = 900,\tilde k_2 = 850$ & $\tilde c_1 = 12,\tilde c_2 = 4.5$ & \textbf{---}\\
			\hline
	\end{tabular}}
	\label{Table-2dof-do}
\end{table}
\noindent For case-I, a 2-DOF system with duffing oscillator attached at both degrees of freedom is considered,
\begin{equation}
\begin{array}{c}
   m_1\ddot x_1+c_1\dot x_1+c_2(\dot x_1-\dot x_2)+k_1x_1+k_2(x_1-x_2)+\alpha_{do}x_1^3+\alpha_{do}(x_1-x_2)^3= f_1+\sigma_1\dot W_1,\\
   m_2\ddot x_2+c_2(\dot x_2-x_1)+k_2(\dot x_2-x_1)+\alpha_{do}(x_2-x_1)^3 = f_2+\sigma_2\dot W_2,
\end{array}
\label{C-I-OS}
\end{equation}
where $m_i$, $c_i$ and $k_i$ are the mass, damping and stiffness respectively for $i-$th degree of freedom.
$f_i$ is deterministic force acting at $i-$th degree of freedom and $\sigma_i$ is the intensity of white noise $\dot W_i$.
$\alpha_{do}$ is the constant for duffing oscillator in Eq. (\ref{C-I-OS}).
For illustrating the proposed approach, we consider the exact form of the governing equation be be a-priori unknown; instead, the governing equation provided takes the following form:
\begin{equation}
\begin{array}{c}
   m_1\ddot x_1+\tilde c_1\dot x_1+\tilde c_2(\dot x_1-\dot x_2)+\tilde k_1x_1+\tilde k_2(x_1-x_2)= f_1+\sigma_1\dot W_1\\
   m_2\ddot x_2+\tilde c_2(\dot x_2-x_1)+\tilde k_2(\dot x_2-x_1) = f_2+\sigma_2\dot W_2,
\end{array}
\label{C-I-AS}
\end{equation}
where $m_i$, $\tilde c_i$ and $\tilde k_i$ are the mass, damping and stiffness respectively for $i-$th degree of freedom. Note that while the actual system in nonlinear, we only have access to a linear system. Additionally, there is slight variation in stiffness and damping as well. Details on the same is shown in Table \ref{Table-2dof-do}.
Naturally not knowing the presence of non-linearity will result in model form error.
The objective here is to identify the model form error by using the known governing equation in Eq. (\ref{C-I-AS}) and noisy measurements.
Note that the identified model form error should be meaningful in the sense that the same can be used for computing responses when the system is subjected to different loading scenario.
For generating data Taylor 1.5\cite{tripura2020ito} strong algorithm has been used and the system is subjected to a realization (deterministic) of frequency restricted (0.5-4Hz) white noise along with the stochastic forces having intensity $\sigma_i = 0.05$. We have considered a sampling frequency of 200 Hz.

\begin{figure}[ht!]
	\begin{subfigure}{1\textwidth}
		\centering
		\includegraphics[scale = 0.25]{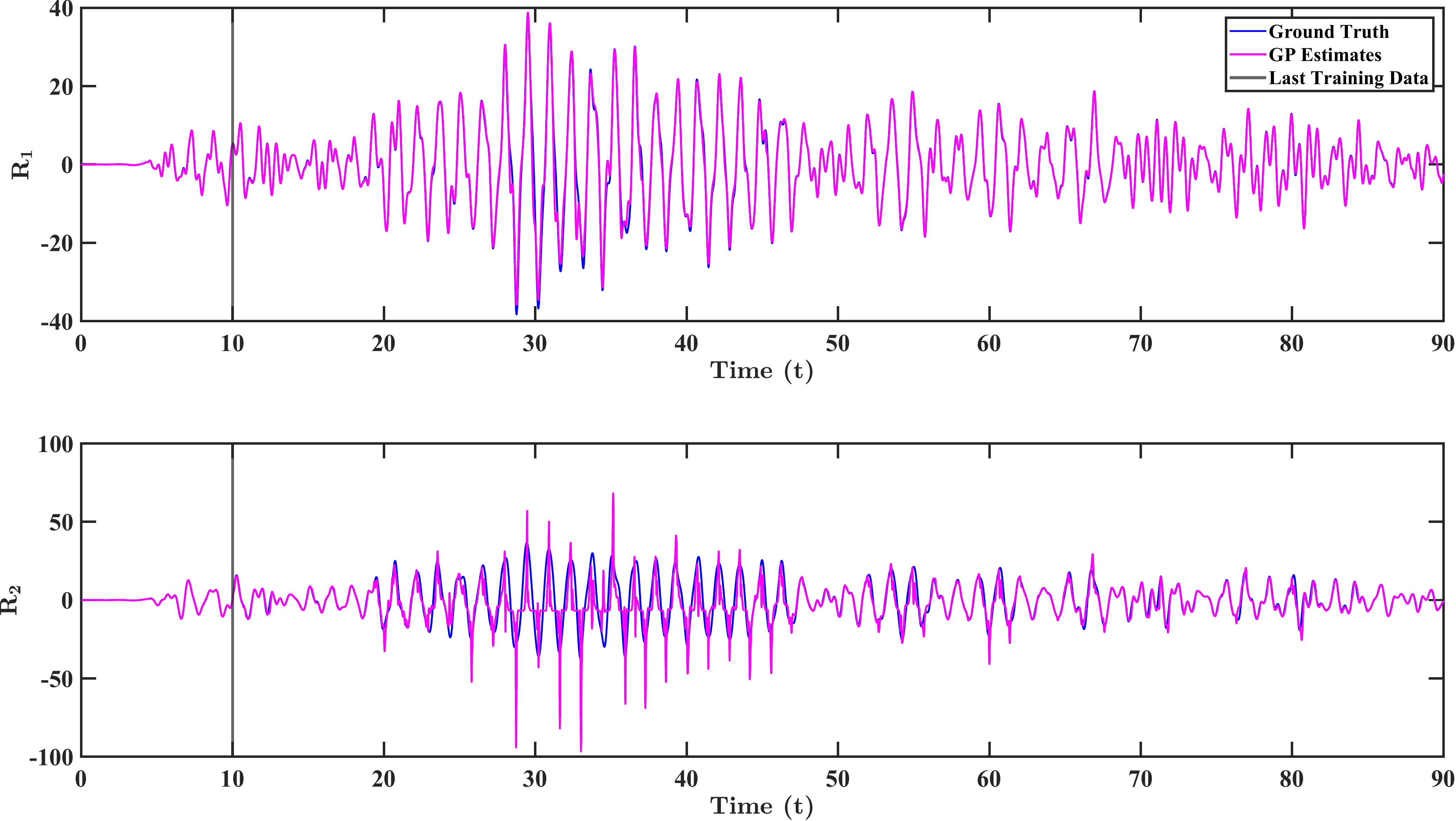}
		\caption{\centering GP estimates compared against ground truth when training data for GPR is provided up-to 20 seconds.}
		\label{2dof-do-gp-1}
	\end{subfigure}
	\begin{subfigure}{1\textwidth}
		\centering
		\includegraphics[scale = 0.25]{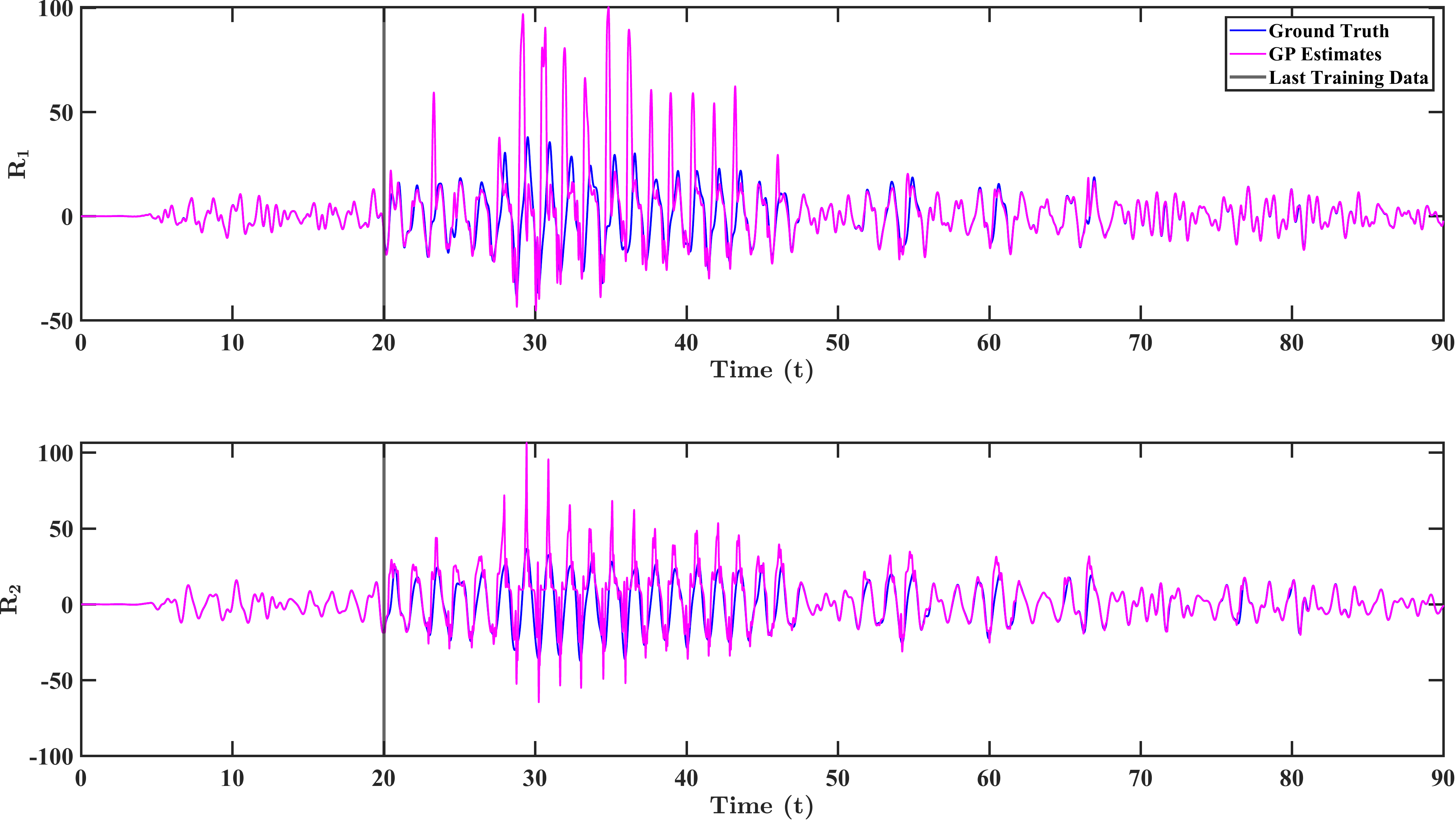}
		\caption{\centering GP estimates compared against ground truth when training data for GPR is provided up-to 30 seconds.}
		\label{2dof-do-gp-2}
	\end{subfigure}
	\begin{subfigure}{1\textwidth}
	\centering
	\includegraphics[scale = 0.25]{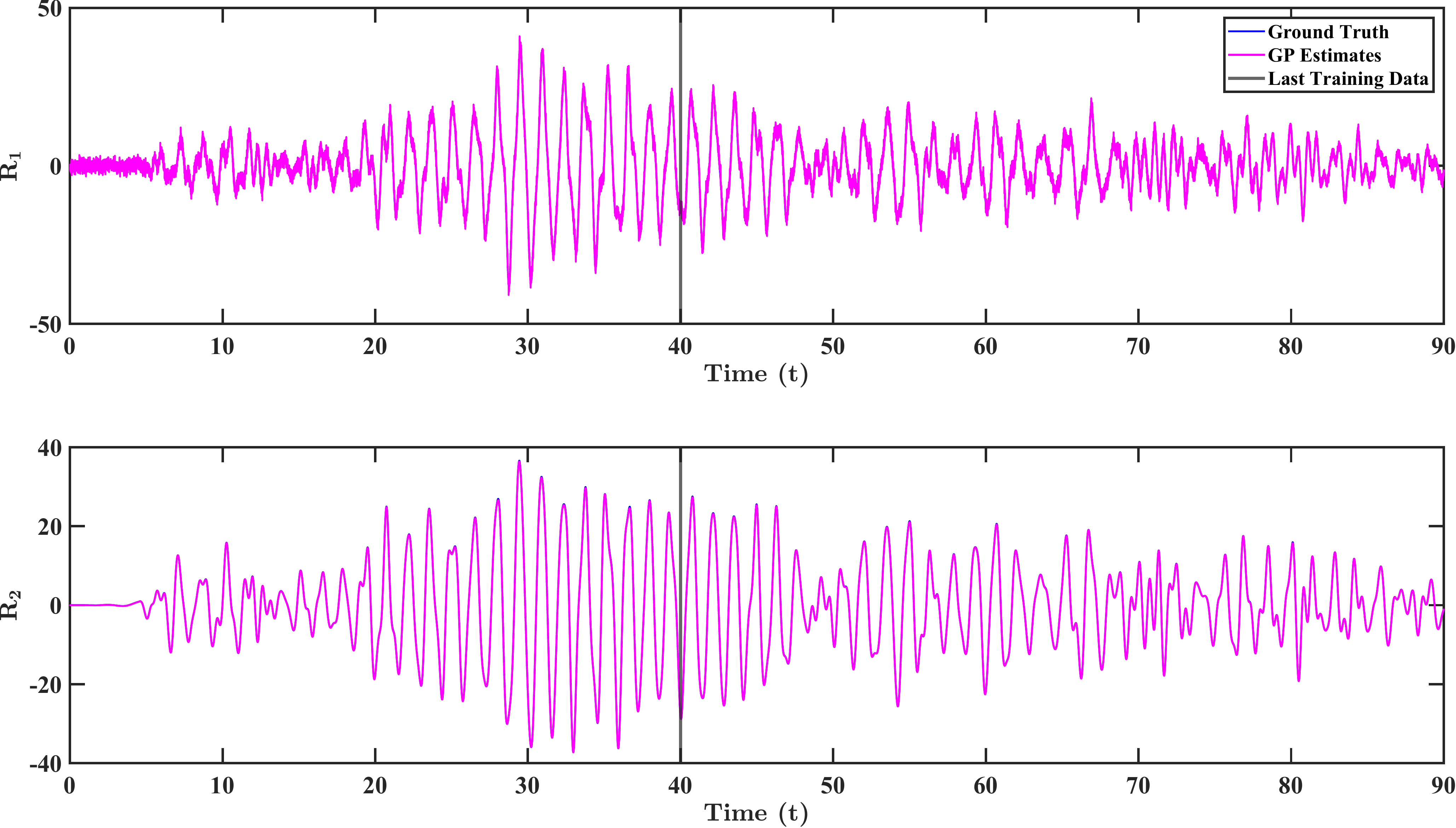}
	\caption{\centering GP estimates compared against ground truth when training data for GPR is provided up-to 40 seconds.}
	\label{2dof-do-gp-3}
\end{subfigure}
	\caption{\centering Projected residual forces (magenta) compared against ground truth (blue) for Case-I}
	\label{2dof-do-gp}
\end{figure}

Fig. \ref{2dof-do-gp} shows the model-form error predicted using the proposed approach. Two cases corresponding to observation time-window of $20$s and $40$s have been considered. We observe that with increase in observation time-window, the proposed approach is able to capture the model form error almost exactly. Interestingly, the proposed approach accurately captures the model-form error until $90$s, which is more than two times the observation window. This illustrates the extrapolation capability of the proposed approach beyond the observation window.

\begin{figure}[ht!]
	\begin{subfigure}{1\textwidth}
		\centering
		\includegraphics[width = 0.85\textwidth]{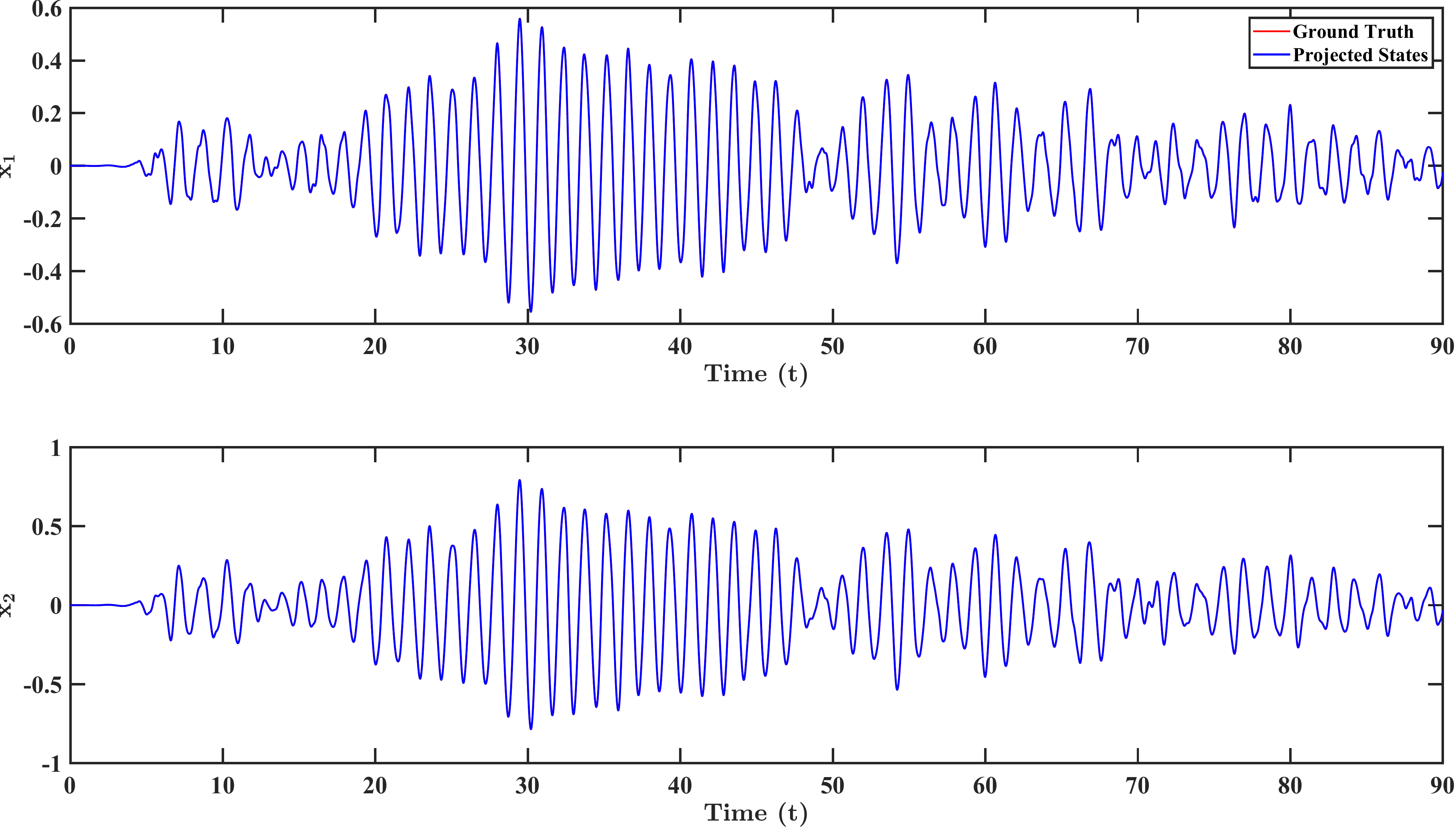}
		\caption{\centering Displacements.}
		\label{2dof-do-sf-1}
	\end{subfigure}
	\begin{subfigure}{1\textwidth}
		\centering
		\includegraphics[width = 0.85\textwidth]{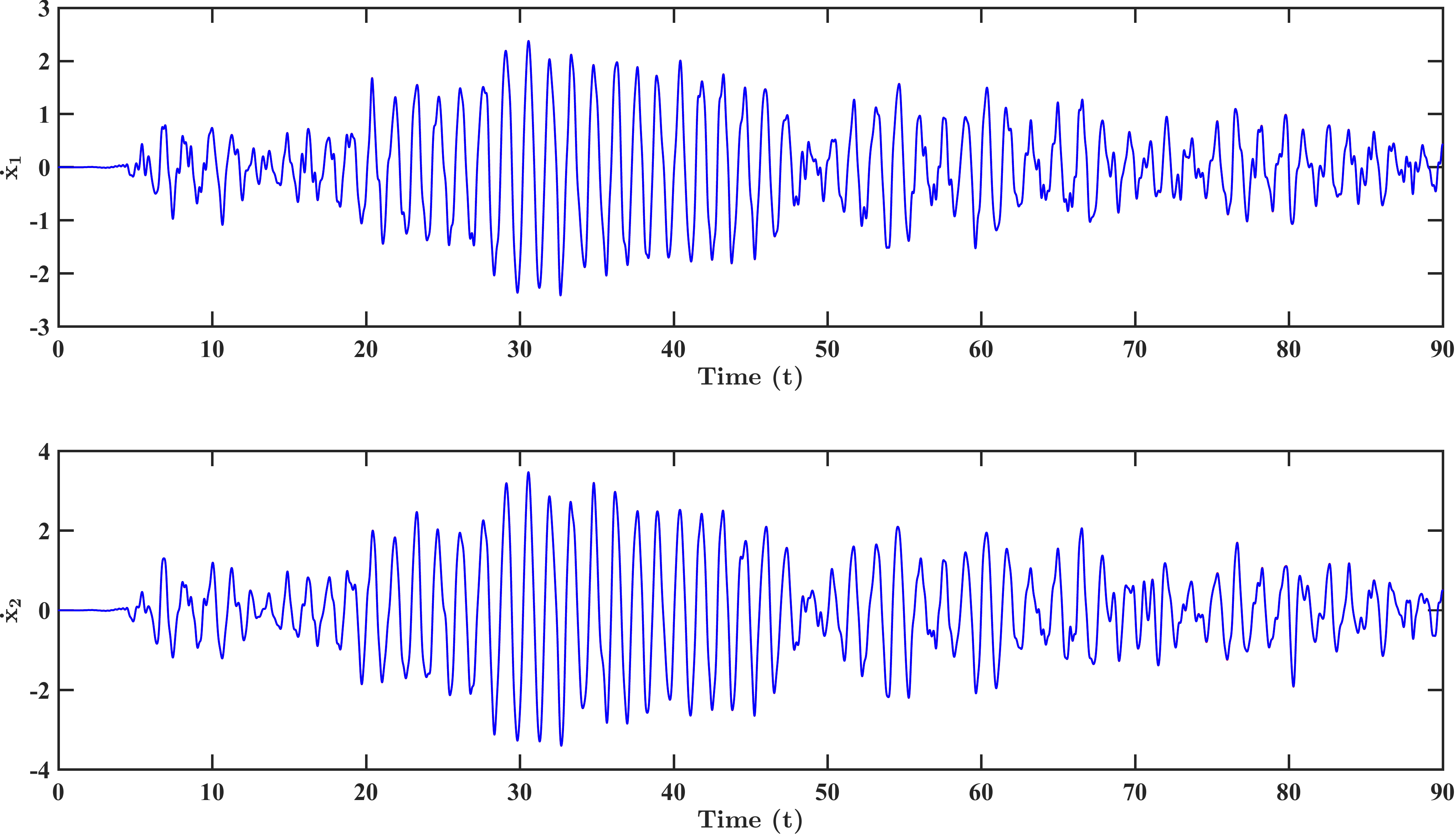}
		\caption{\centering Velocities.}
		\label{2dof-do-sf-2}
	\end{subfigure}
	\caption{\centering Projected states (blue) compared against ground truth (red) when system is subjected to 'same input' for case-I}
	\label{2dof-do-sf}
\end{figure}

\begin{figure}[ht!]
	\begin{subfigure}{1\textwidth}
		\centering
		\includegraphics[width = 0.85\textwidth]{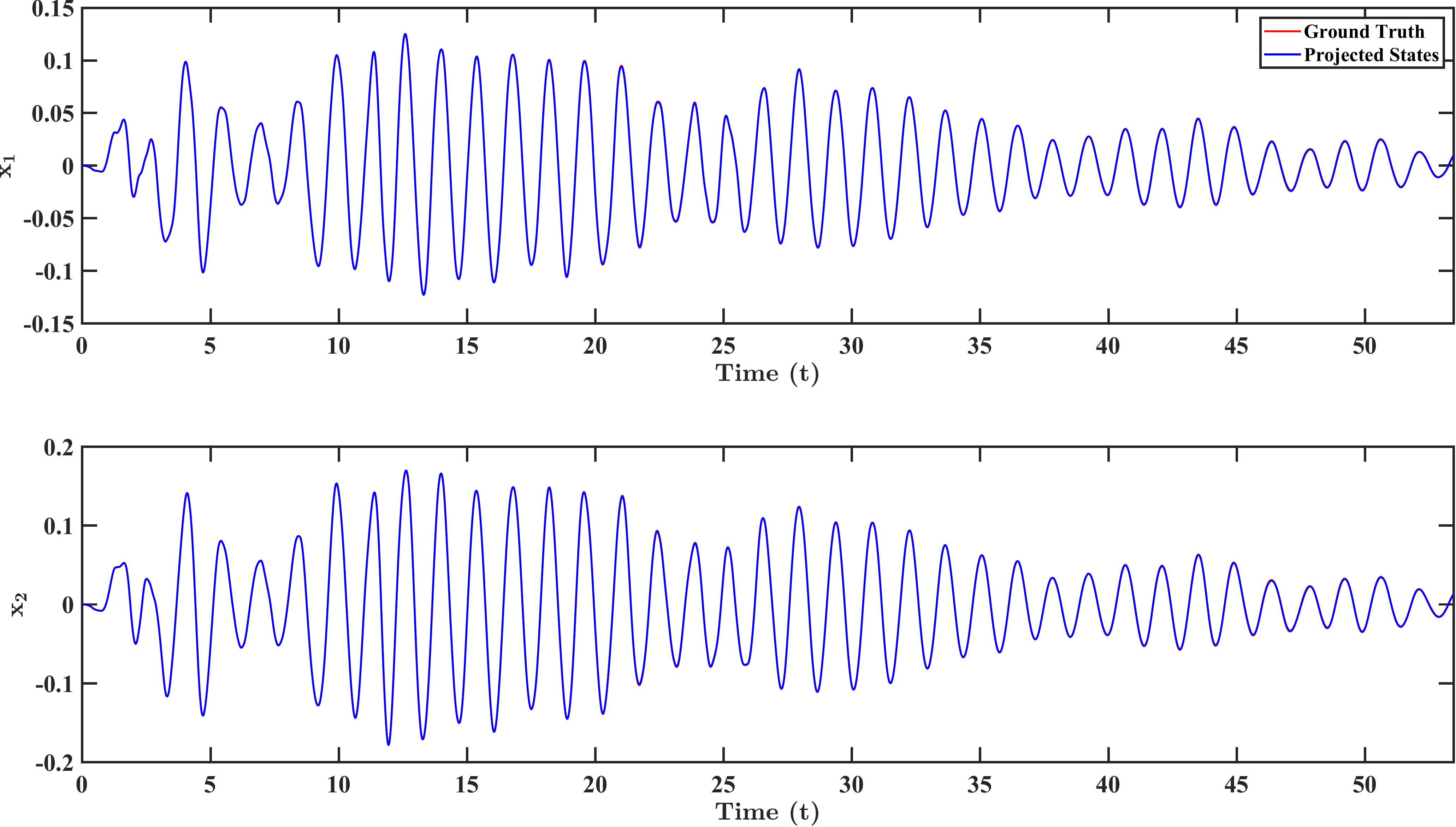}
		\caption{\centering Displacements.}
		\label{2dof-do-df-1}
	\end{subfigure}
	\begin{subfigure}{1\textwidth}
		\centering
		\includegraphics[width = 0.85\textwidth]{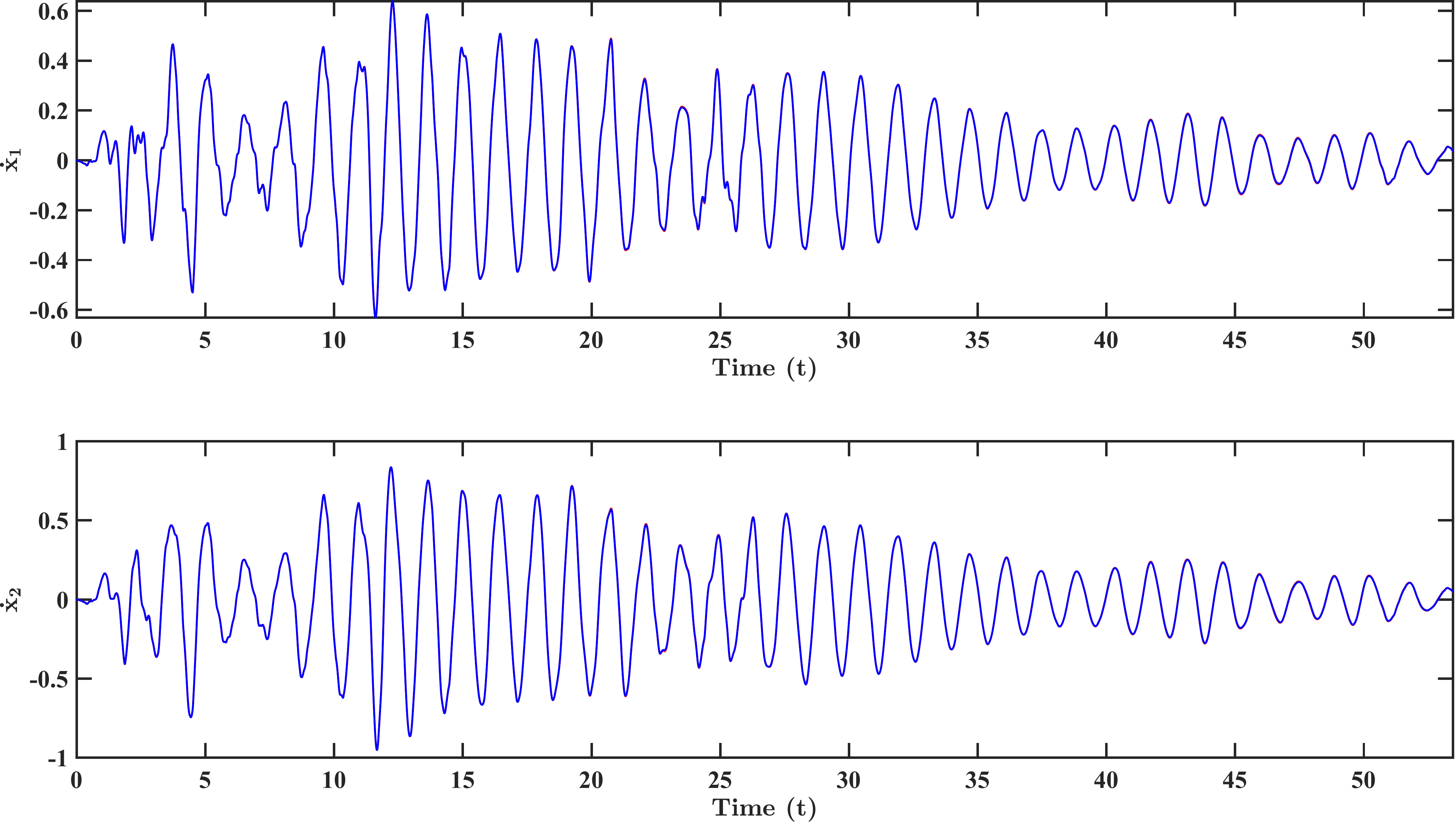}
		\caption{\centering Velocities.}
		\label{2dof-do-df-2}
	\end{subfigure}
	\caption{\centering Projected states (blue) compared against ground truth (red) when system is subjected to 'different input' for case-I}
	\label{2dof-do-df}
\end{figure}

Having showcased the excellent performance of the proposed approach in identifying the model-form, we proceed to estimating the responses of the underlying system. To that end, we include the model-form error model (represented in term of GP) into the governing equation as an additional term and solve the forward problem. Given the fact that the model-form error for the $40$s observation-window is better, we utilize the same in this case. Fig. \ref{2dof-do-sf} shows the results when forward problem is solved using the same input as that used for filtering.
We observe that even without knowing the duffing oscillator parameters or the exact nature of non-linearity, we can reliably estimate the states of original system.
To illustrate the generalization of the proposed approach to unseen environment, we consider a case where the system is subjected to Imperial Valley: El-Centro Earthquake ground motion data. We note that the model has not seen this motion during the training phase. The results obtained for this case are shown in Fig. \ref{2dof-do-df}. We observe that the responses predicted using the proposed approach matches almost exactly with the ground truth. This indicate that the proposed approach approach is able to generalize to unseen environment.
\subsection{Case-II : MDOF system with Bouc-Wen oscillator}
\noindent As the second example, we consider MDOF systems with Bouc-Wen oscillator fixed to the first DOF. The governing equation for this system is represented as:
\begin{equation}
  \begin{array}{c}
  m_1\ddot x_1+c_1\dot x_1+c_2(\dot x_1-\dot x_2)+k_1x_1+k_2(x_1-x_2)+(1-k_r)Q_y\,z= f_1+\sigma_1\dot W_1\\
   m_2\ddot x_2+c_2(\dot x_2-x_1)+k_2(\dot x_2-x_1) = f_2+\sigma_2\dot W_2\\
   \dot z = \frac{1}{D_y}(\alpha_{bw} \dot x_1-\gamma z |\dot x_1||z|^{\eta-1}-\beta \dot x_1 |z|^\eta),
  \end{array}
\end{equation}
where $D_y$, $\alpha_{bw}$, $\beta$, $\gamma$, $\eta$, $k_r$ and $Q_y$ are parameters specific to Bouc-Wen oscillator. Reader can read more about Bouc-Wen system here \cite{shirali2009principal}. Similar to the previous example, we consider the exact governing equation to be unknown; instead, the known governing equation is linear in nature. Additionally, exact system parameters of the underlying system are also known in an approximate sense only.
System parameters for the original and known systems are given in Table \ref{Table-2dof-bw}. The objective here is to estimate the model-form error and use the same to update the known but approximate system. Similar to previous case, synthetic data is generated by subjecting the system to a realization (deterministic) of frequency restricted (0.5-4Hz) white noise along with the stochastic forces having intensity $\sigma_i = 0.01$, and analysing it using Taylor 1.5 strong algorithm.

\begin{table}[ht!]
\caption{System parameters for 2-DOF Bouc-Wen example.}
	\label{Table-2dof-bw}
	\centering\scalebox{0.75}{
		\begin{tabular}{|c|c|c|c|c|}
			\hline
			System & Mass (Kg) & Stiffness (N/m) & Damping (Ns/m) & Non-linear Parameters \\ \hline
			Original &$m_1 = 5,m_2 = 20$ & $k_1 = 1000,k_2 = 2000$ & $c_1 = 7.5,c_2 = 20$ & $\begin{matrix}Q_y = 0.05\sum\limits_{i} m_ig, k_r = \frac{1}{6},\\ \alpha_{bw} = 1, \beta_{bw} = 0.5, \gamma = 0.5, D_y = 0.013, \end{matrix}$ \\
			Known &$m_1 = 5,m_2 = 20$& $\tilde k_1 = 900,\tilde k_2 = 850$ & $\tilde c_1 = 12,\tilde c_2 = 4.5$ & \textbf{---}
			\\\hline
	\end{tabular}}
\end{table}

\begin{figure}[ht!]
	\begin{subfigure}{1\textwidth}
		\centering
		\includegraphics[width = 0.85\textwidth]{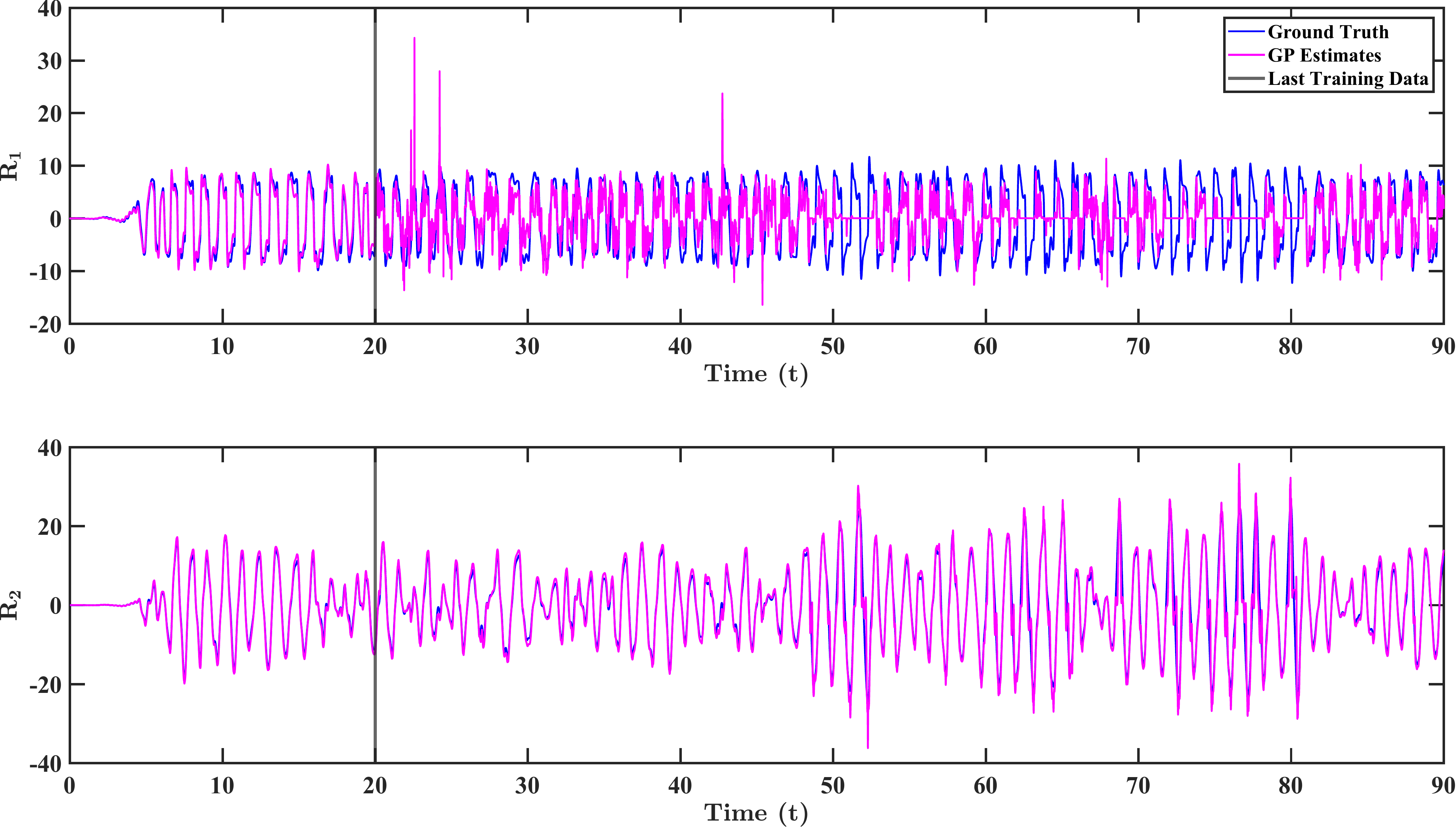}
		\caption{\centering GP estimates compared against ground truth when training data for GPR is provided up-to 20 seconds.}
		\label{2dof-bw-gp-1}
	\end{subfigure}
	\begin{subfigure}{1\textwidth}
		\centering
		\includegraphics[width = 0.85\textwidth]{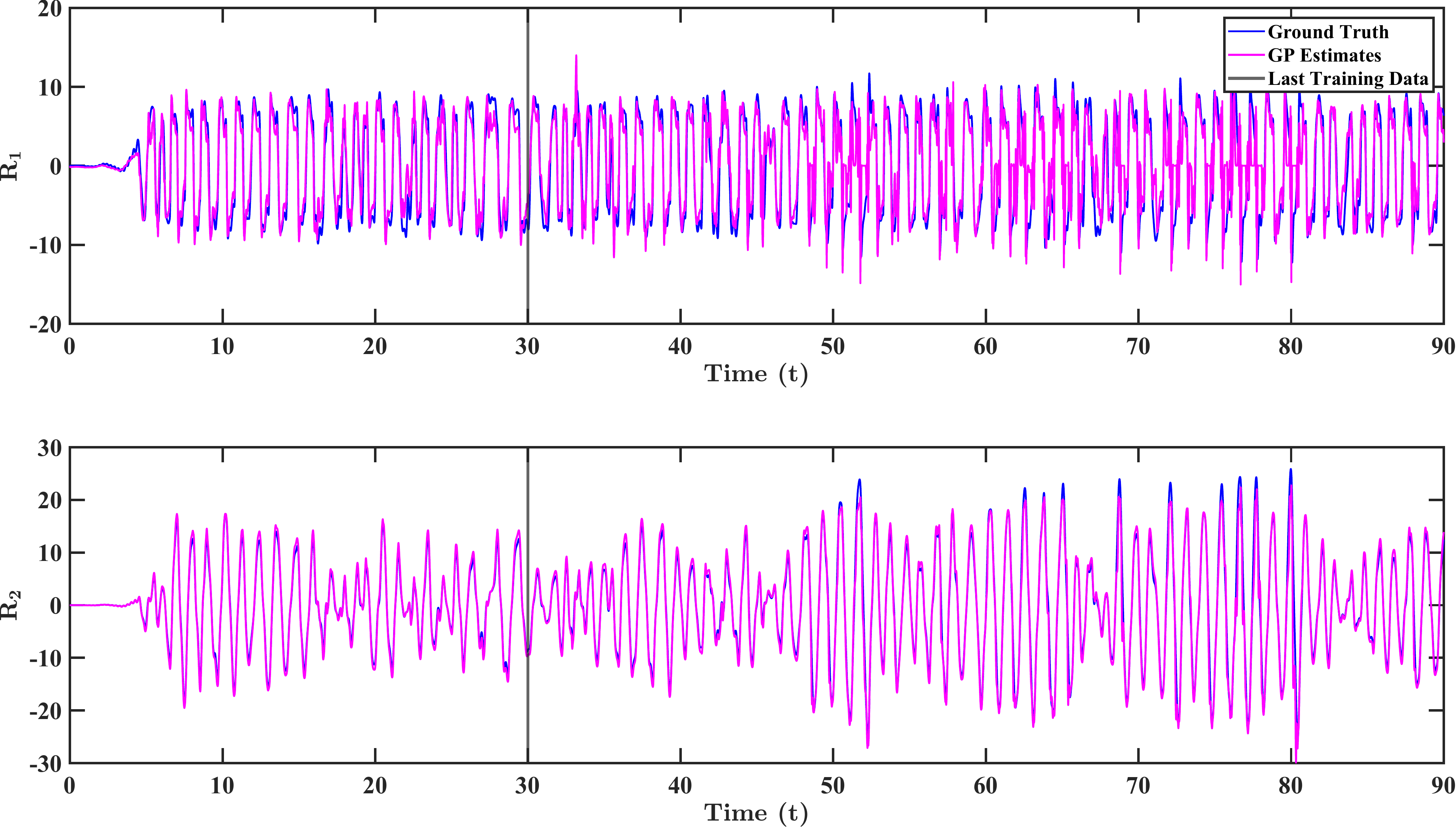}
		\caption{\centering GP estimates compared against ground truth when training data for GPR is provided up-to 30 seconds.}
		\label{2dof-bw-gp-2}
	\end{subfigure}
	\caption{\centering Projected residual forces (magenta) compared against ground truth (blue) for 2-DOF system with Bouc-Wen oscillator fixed to the first DOF.}
	\label{2dof-bw-gp}
\end{figure}

We employ the proposed framework to estimate the model-form error by using the simulated data and the known but approximate governing equation. Similar to previous example, we illustrate the performance by taking observation time-window of $20$s and $30$s. Fig. \ref{2dof-bw-gp} shows the results corresponding to the two cases. As expected, the results produced with longer observation window better represent the ground truth. Again, the proposed approach is able to identify the model-form error up to $90$s, which is three time the observation window. This illustrates the capability of the proposed approach in generalizing beyond the observation window.

\begin{figure}[ht!]
	\begin{subfigure}{1\textwidth}
		\centering
		\includegraphics[width = 0.85\textwidth]{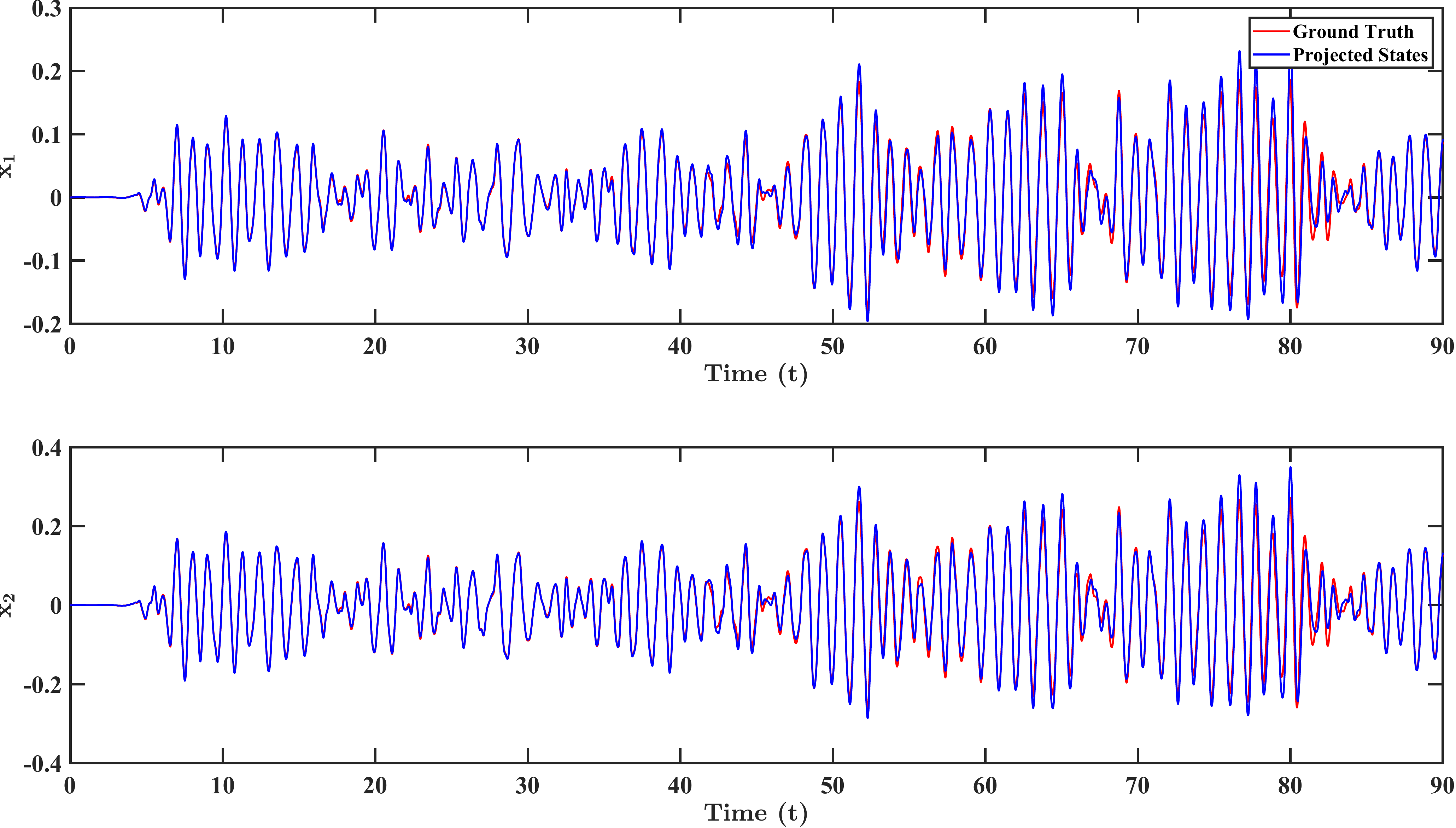}
		\caption{\centering Displacements.}
		\label{2dof-bw-sf-1}
	\end{subfigure}
	\begin{subfigure}{1\textwidth}
		\centering
		\includegraphics[width = 0.85\textwidth]{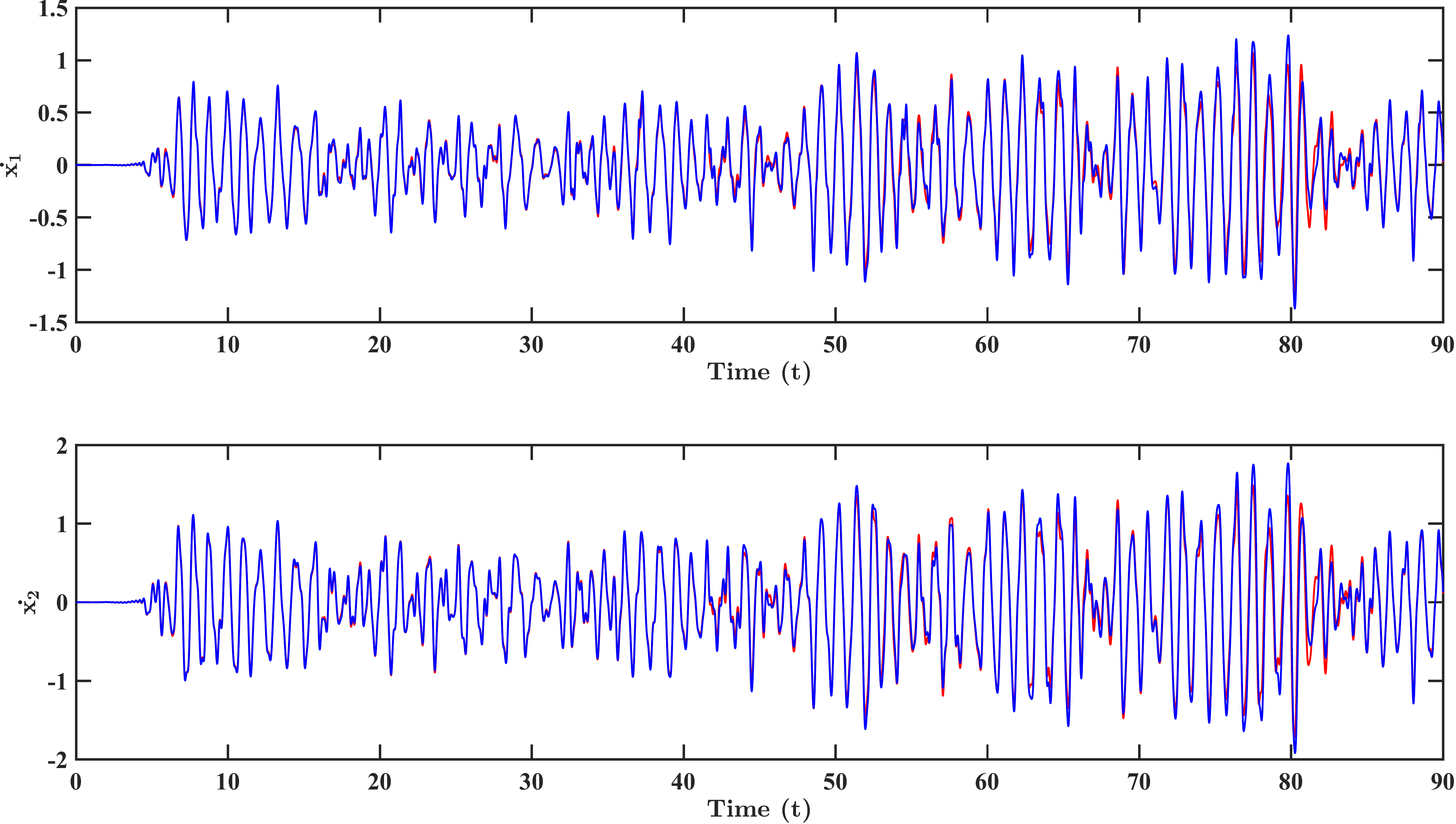}
		\caption{\centering Velocities.}
		\label{2dof-bw-sf-2}
	\end{subfigure}
	\caption{\centering Projected states (blue) compared against ground truth (red) when system is subjected to 'same input' for 2-DOF Bouc-Wen example}
	\label{2dof-bw-sf}
\end{figure}

\begin{figure}[ht!]
	\begin{subfigure}{1\textwidth}
		\centering
		\includegraphics[width = 0.85\textwidth]{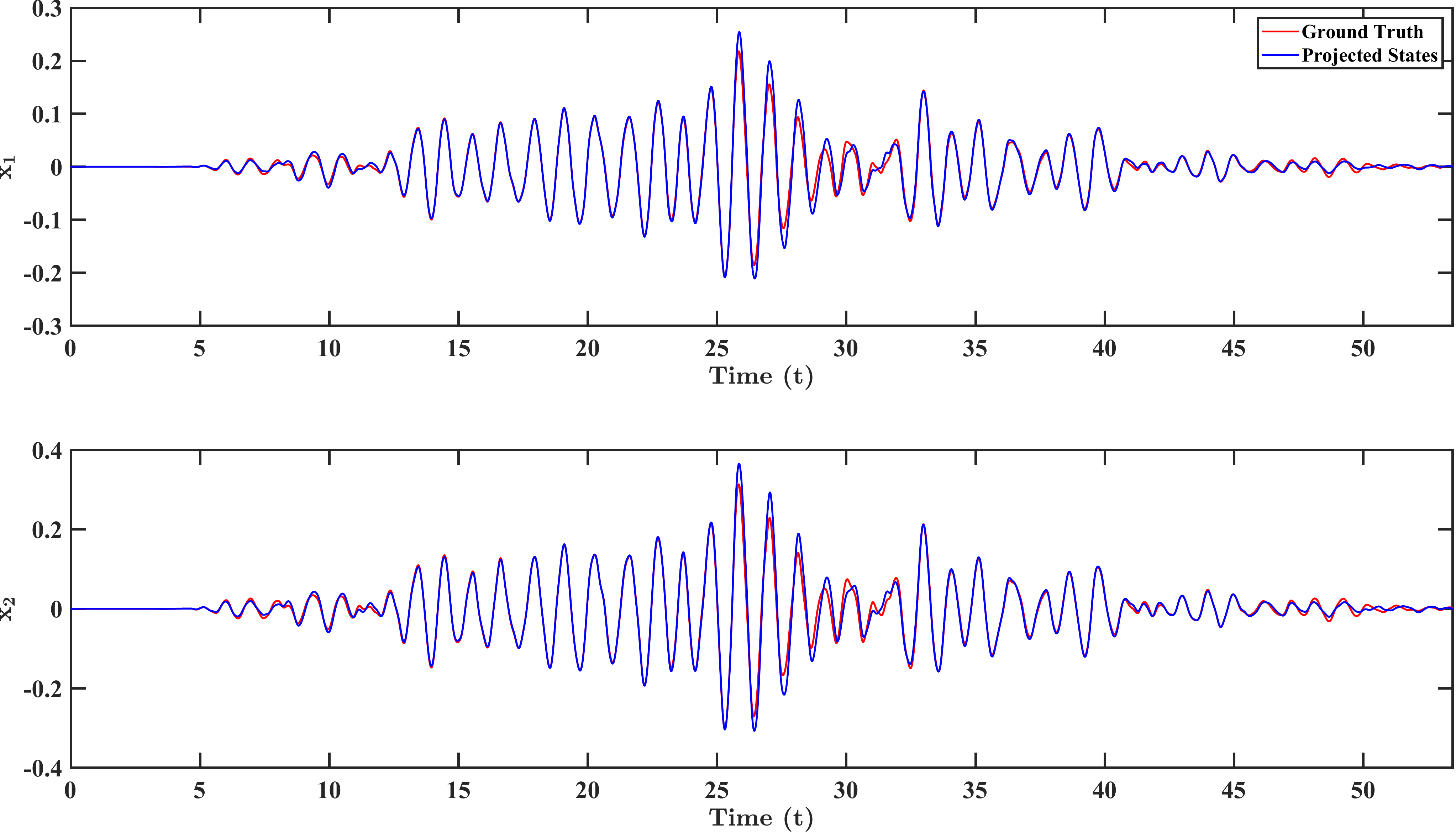}
		\caption{\centering Displacements.}
		\label{2dof-bw-df-1}
	\end{subfigure}
	\begin{subfigure}{1\textwidth}
		\centering
		\includegraphics[width = 0.85\textwidth]{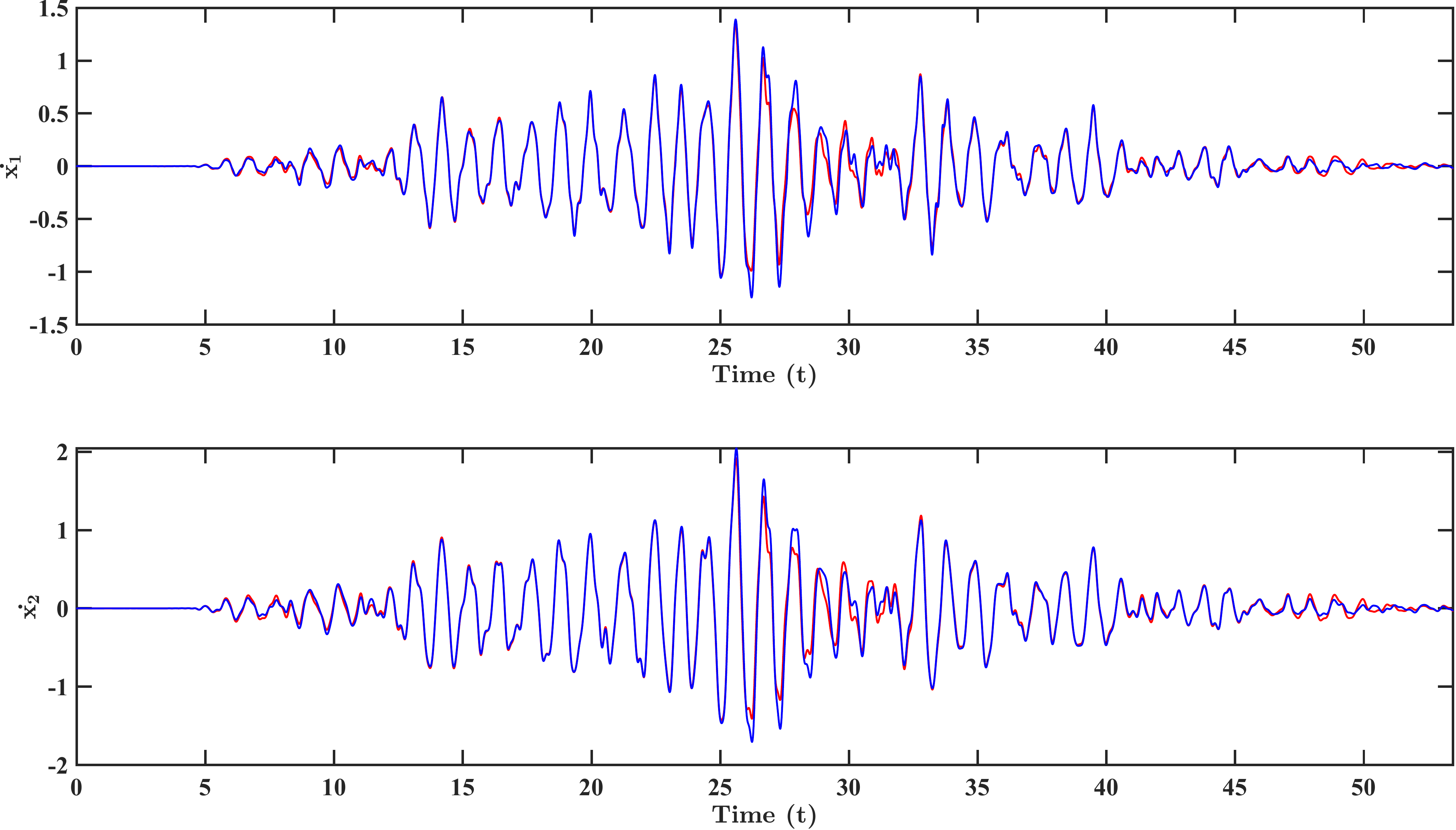}
		\caption{\centering Velocities.}
		\label{2dof-bw-df-2}
	\end{subfigure}
	\caption{\centering Projected states (blue) compared against ground truth (red) when system is subjected to 'different input' for 2-DOF Bouc-Wen example}
	\label{2dof-bw-df}
\end{figure}

Next, we proceed to examine the performance of the proposed approach in predicting the systems response. To that end, the identified model-form error is included into the governing equation as an additional term. Fig. \ref{2dof-bw-sf} shows results obtained using the proposed approach. For this case, same input as that used for training is used. We observe that the projected states closely follow the ground truth. To illustrate the ability of the proposed model to generalize to new environment, we subject the system to an unseen input. For this example, the unseen input is a realization of frequency restricted (0.5Hz - 4Hz), amplitude modulated (Hamming window) white noise. The velocity and displacement time history obtained using the proposed approach are shown in Fig.  \ref{2dof-bw-df}. We observe that the proposed approach is able to accurately predict the responses. 

Finally, to illustrate the scalability of the proposed approach, we consider a case where the underlying system has five degrees of freedom, with Bouc-Wen oscillator connected to the first DOF.The governing equation for this case is as follows:
\begin{equation}
  \begin{array}{c}
  m_1\ddot x_1+c_1\dot x_1+c_2(\dot x_1-\dot x_2)+k_1x_1+k_2(x_1-x_2)+(1-k_r)Q_y\,z= f_1+\sigma_1\dot W_1\\
  m_2\ddot x_2+c_2(\dot x_2-x_1)+c_3(\dot x_2-x_3)+k_2(\dot x_2-x_1)+k_3(\dot x_2-x_3) = f_2+\sigma_2\dot W_2\\
  m_3\ddot x_3+c_3(\dot x_3-x_2)+c_4(\dot x_3-x_4)+k_3(\dot x_3-x_2)+k_4(\dot x_3-x_4) = f_3+\sigma_3\dot W_3\\
  m_4\ddot x_4+c_4(\dot x_4-x_3)+c_5(\dot x_4-x_5)+k_4(\dot x_4-x_3)+k_5(\dot x_4-x_5) = f_4+\sigma_4\dot W_4\\
  m_5\ddot x_5+c_5(\dot x_5-x_4)+k_5(\dot x_5-x_4) = f_5+\sigma_5\dot W_5\\
  \dot z = \frac{1}{D_y}(\alpha_{bw} \dot x_1-\gamma z |\dot x_1||z|^{\eta-1}-\beta \dot x_1 |z|^\eta)
  \end{array}.
\end{equation}
The original equation is not available a-priori; instead, we have been provided the following linear equations,
\begin{equation}
  \begin{array}{c}
  m_1\ddot x_1+\tilde c_1\dot x_1+\tilde c_2(\dot x_1-\dot x_2)+\tilde k_1x_1+\tilde k_2(x_1-x_2)+R_1= f_1+\sigma_1\dot W_1\\
  m_2\ddot x_2+\tilde c_2(\dot x_2-x_1)+\tilde c_3(\dot x_2-x_3)+\tilde k_2(\dot x_2-x_1)+\tilde k_3(\dot x_2-x_3)+R_2 = f_2+\sigma_2\dot W_2\\
  m_3\ddot x_3+\tilde c_3(\dot x_3-x_2)+\tilde c_4(\dot x_3-x_4)+\tilde k_3(\dot x_3-x_2)+\tilde k_4(\dot x_3-x_4)+R_3 = f_3+\sigma_3\dot W_3\\
  m_4\ddot x_4+\tilde c_4(\dot x_4-x_3)+\tilde c_5(\dot x_4-x_5)+\tilde k_4(\dot x_4-x_3)+\tilde k_5(\dot x_4-x_5)+R_4 = f_4+\sigma_4\dot W_4\\
  m_5\ddot x_5+\tilde c_5(\dot x_5-x_4)+\tilde k_5(\dot x_5-x_4)+R_5 = f_5+\sigma_5\dot W_5\\
  \end{array}
  \label{5dof-linear}
\end{equation}
Again, the parameters of the known (approximate) and the original systems are not identical. Details on the same are given in Table \ref{Table-5dof-bw}. Training data for this case is generated using Taylor 1.5 strong scheme. In this case, we have considered a sampling frequency of 1000 Hz. Other settings are kept same as before.

\begin{table}[ht!]
\caption{System parameters for 5-DOF Bouc-Wen example.}
	\centering\scalebox{0.8}{
		\begin{tabular}{|c|c|c|c|c|}
			\hline
			System & Mass (Kg) & Stiffness (N/m) & Damping (Ns/m) & Non-linear Parameters \\ \hline
			Original & $\begin{matrix}m_1 = 400, m_2 = 380\\m_3 = 360, m_4 = 340\\m_5 = 320\end{matrix}$ & $\begin{matrix}k_1 = 100000,k_2 = 200000\\k_3 = 190000,k_4 = 180000\\k_5 = 170000\end{matrix}$ & $\begin{matrix}c_1 = 100,c_2 = 200\\c_3 = 190,c_4 = 180\\c_5 = 170\end{matrix}$ & $\begin{matrix}Q_y = 0.05\sum\limits_{i} m_ig\\ k_r = \frac{1}{6}, \alpha_{bw} = 1, \beta_{bw} = 0.5,\\\gamma = 0.5, D_y = 0.013, \end{matrix}$ \\\hline
			Known &$\begin{matrix}m_1 = 400, m_2 = 380\\m_3 = 360, m_4 = 340\\m_5 = 320\end{matrix}$& $\begin{matrix}\tilde k_1 = 105000,\tilde k_2 = 210000\\\tilde k_3 = 180500,\tilde k_4 = 171000\\\tilde k_5 = 161500\end{matrix}$ & $\begin{matrix}\tilde c_1 = 110,\tilde c_2 = 210\\\tilde c_3 = 171,\tilde c_4 = 198\\\tilde c_5 = 161.5\end{matrix}$ & \textbf{---}
			\\\hline
	\end{tabular}}
	\label{Table-5dof-bw}
\end{table}

\begin{figure}[ht!]
	\begin{subfigure}{1\textwidth}
		\centering
		\includegraphics[width = 0.85\textwidth]{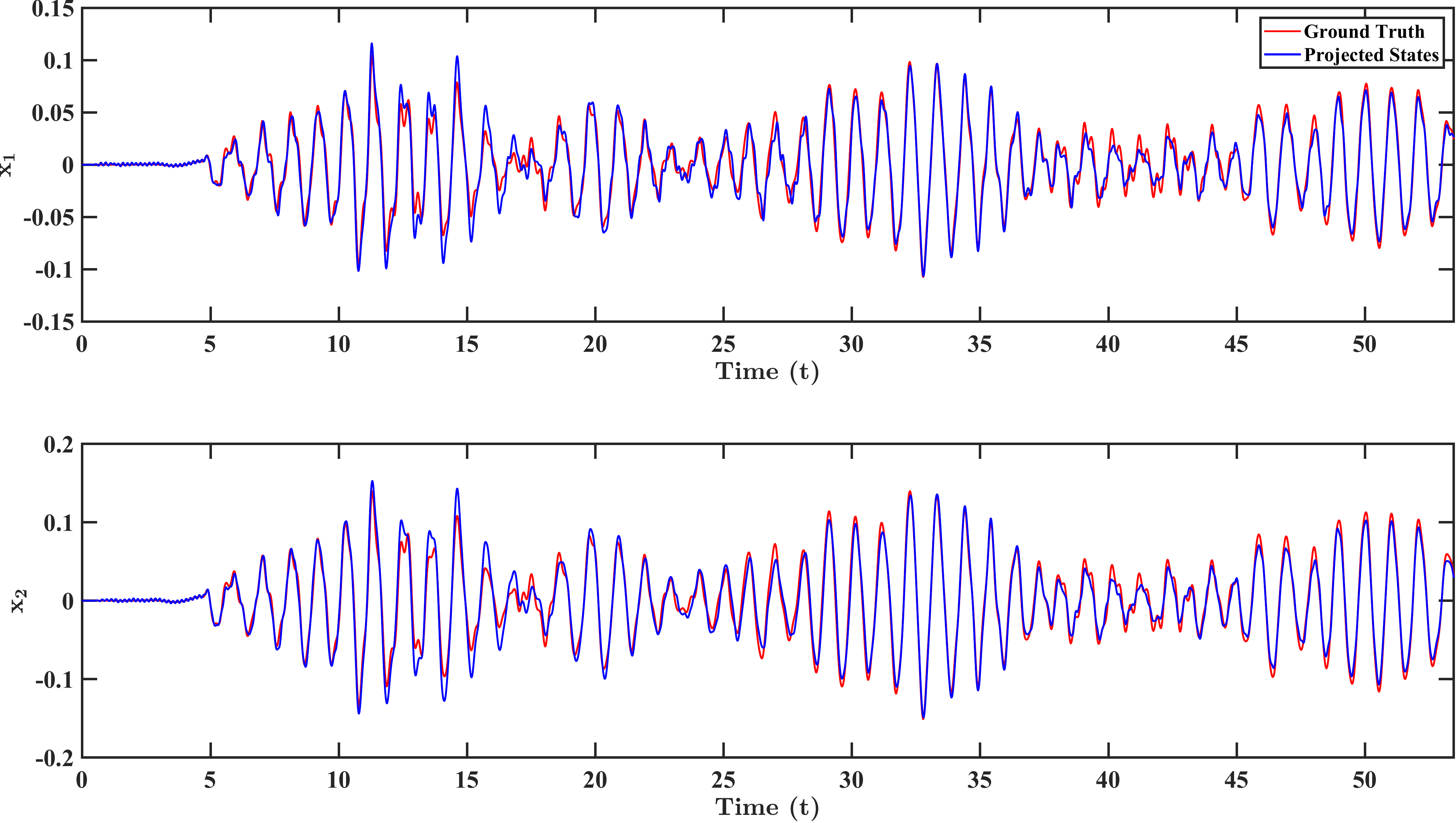}
		\caption*{\tiny \,\,}
		\label{5dof-bw-sf-1}
	\end{subfigure}
	\begin{subfigure}{1\textwidth}
		\centering
		\includegraphics[width = 0.85\textwidth]{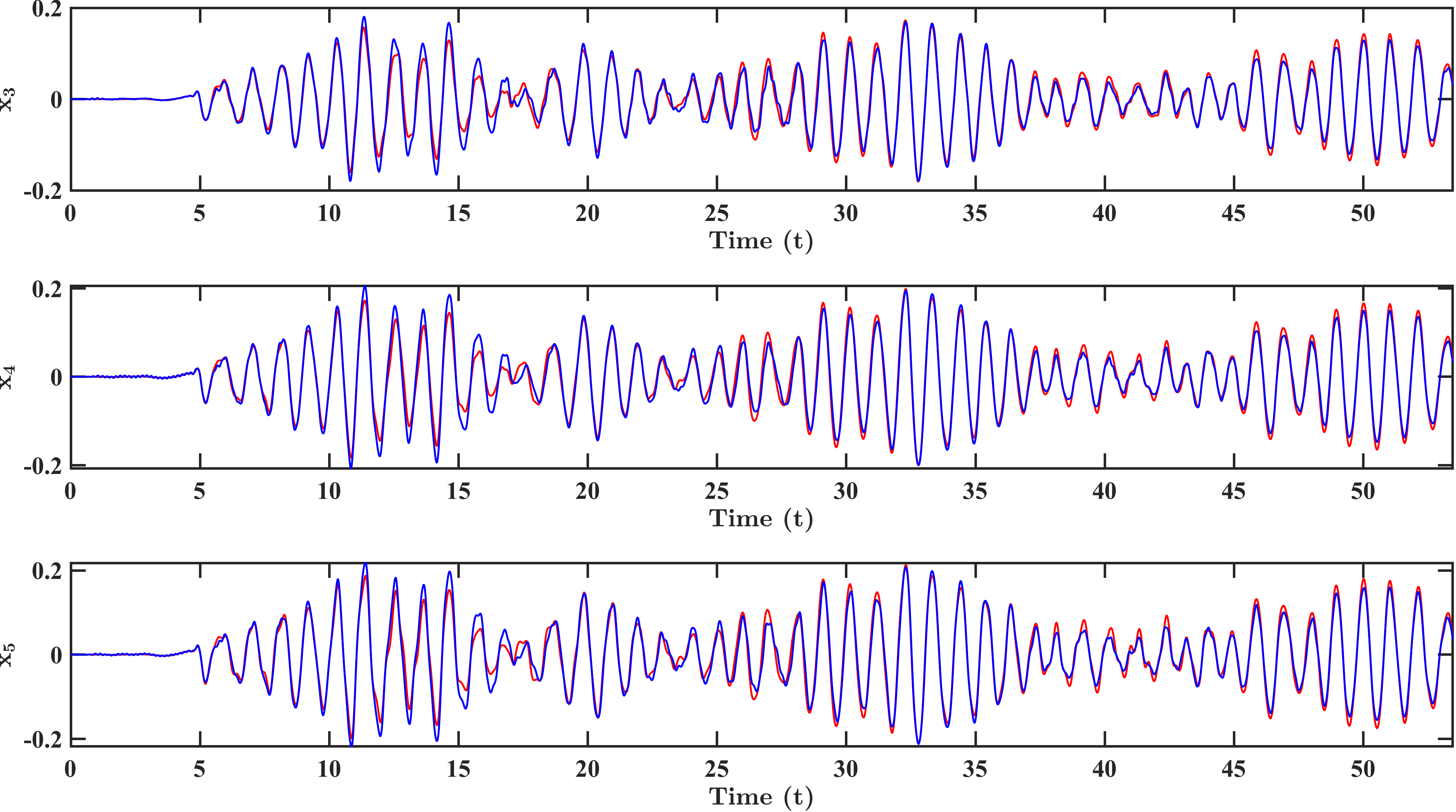}
		\label{5dof-bw-sf-2}
	\end{subfigure}
	\caption{\centering Projected states (blue) compared against ground truth (red) when system is subjected to `same input' for 5-DOF Bouc-Wen example.}
	\label{5dof-bw-sf}
\end{figure}

In this case, we directly proceed to examining the predictive capability of the proposed approach. Fig. \ref{5dof-bw-sf} shows the displacement estimates corresponding to the same input. Excellent match between the predicted displacement and the ground truth is observed. To illustrate the capability of the proposed framework to unseen environment, we also predicted the displacement corresponding to an unseen input (a frequency restricted realization of white noise with frequency ranging between 0.5 Hz to 4 Hz). Results for this case are shown in Fig. \ref{5dof-bw-sf}. In this case also, reasonably good match between the estimated displacement and the ground truth is observed.

\begin{figure}[ht!]
	\begin{subfigure}{1\textwidth}
		\centering
		\includegraphics[width = 0.85\textwidth]{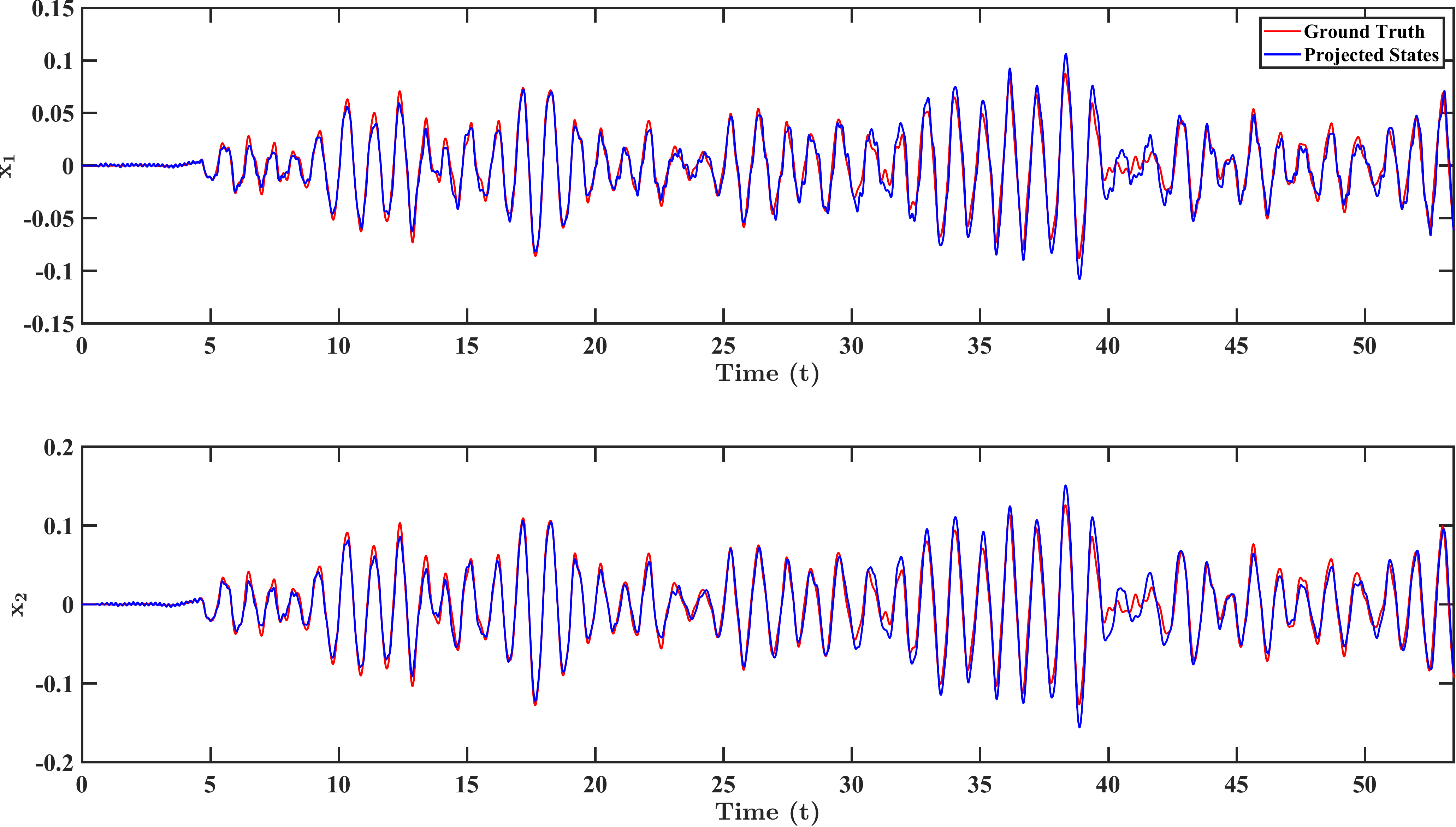}
		\caption*{\tiny \,\,}
		\label{5dof-bw-df-d-1}
	\end{subfigure}
	\begin{subfigure}{1\textwidth}
		\centering
		\includegraphics[width = 0.85\textwidth]{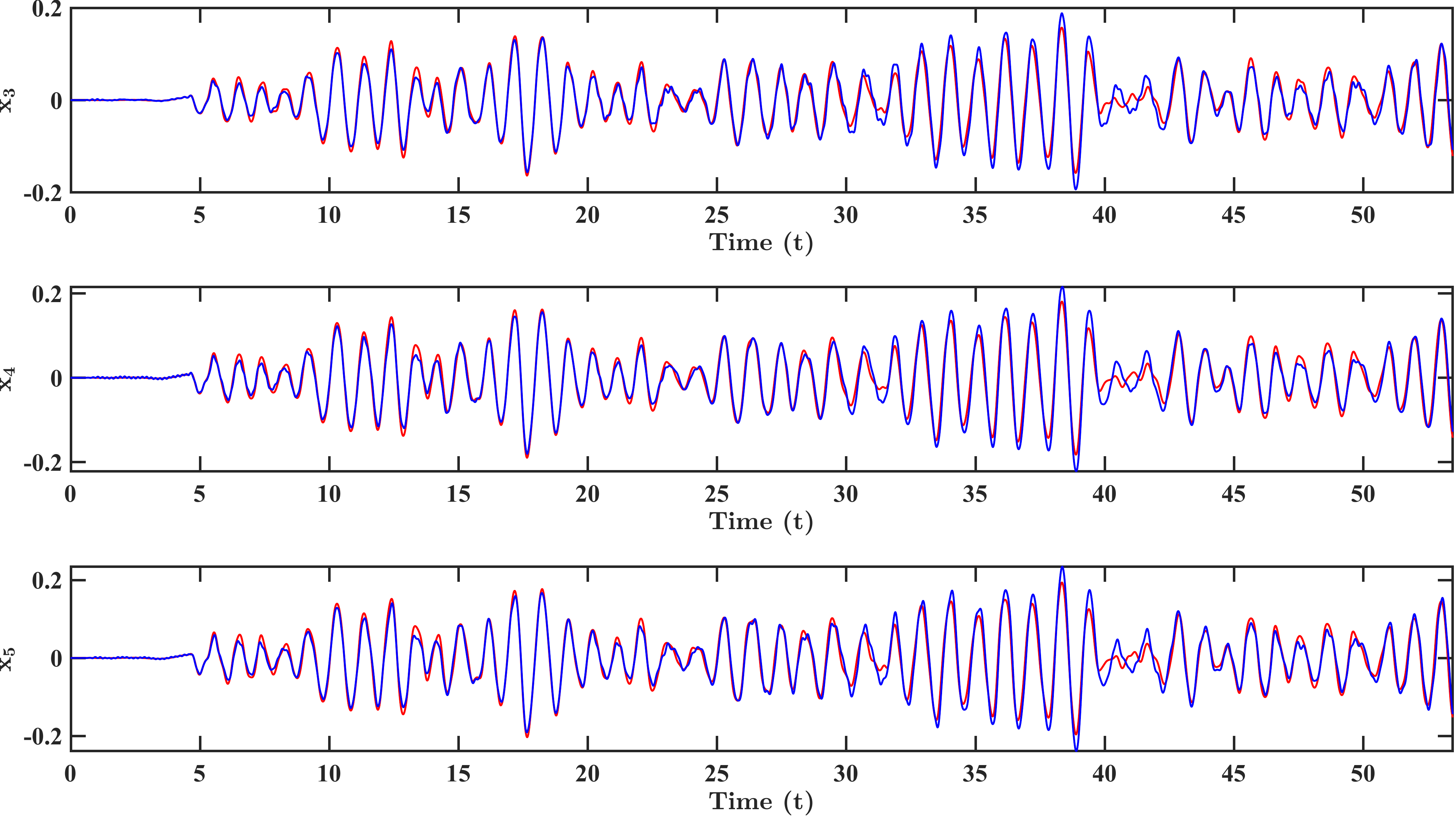}
		\label{5dof-bw-df-d-2}
	\end{subfigure}
	\caption{\centering Projected displacements (blue) compared against ground truth (red) when system is subjected to `different input' for 5-DOF Bouc-Wen example.}
	\label{5dof-bw-df-d}
\end{figure}

\subsection{Case-III : SDOF duffing Van-der Pol oscillator}
\noindent As the last example, we consider a SDOF duffing Van-der Pol (VPD) oscillator \cite{tripura2020ito}. The governing equation for VPD oscillator is given as follows
\begin{equation}
  m\ddot x+c\dot x-kx+\alpha_{dvp}x^3= f+\sigma x\dot W,
\end{equation}
where $\alpha_{dvp}$ is the constant for DVP oscillator. However, the governing equation for DVP is not known apiori; instead, the known governing equation takes the following form:
\begin{equation}
  m\ddot x+c\dot x+kx+\alpha_{do}x^3+R= f+\sigma\dot W.
\end{equation}
Additionally, the system parameters are also known only in an approximate manner. Details on the system parameters are provided in Table \ref{Table-sdof-dvp}. The objective is to identify the model-form error arising due to the difference between the actual and the known systems. Although this is a relatively simpler system, the difficulty arises from the fact that the known system is nonlinear in nature and hence, estimating the model-form error (represented as residual force) becomes challenging. As stated before, we use DUKF for joint input-state estimation in this case. For data generation, the system is subjected to a sinusoidal wave with frequency of 1.59 Hz. The intensity for white noise is taken as $\sigma = 0.10$. 
Taylor 1.5 Strong algorithm has been used for generating synthetic data. 
While testing for different input, we consider a sinusoidal wave with frequency of 2.39 Hz.

\begin{table}[ht!]
\caption{System parameters for Case-III.}
	\centering\scalebox{0.75}{
		\begin{tabular}{|c|c|c|c|c|}
			\hline
			System & Mass (Kg) & Stiffness (N/m) & Damping (Ns/m) & Non-linear Parameters \\\hline 
			Original & $m = 10$ & $k = 100$ & $c = 2.5$ & $\alpha_{dvp} = 10$ \\
			Known & $m = 10$ & $\tilde k = 50$ & $\tilde c = 2.5$ & $\alpha_{do} = 11$
			\\\hline
	\end{tabular}}
	\label{Table-sdof-dvp}
\end{table}

\begin{figure}[ht!]
	\centering
	\includegraphics[width = 1\textwidth]{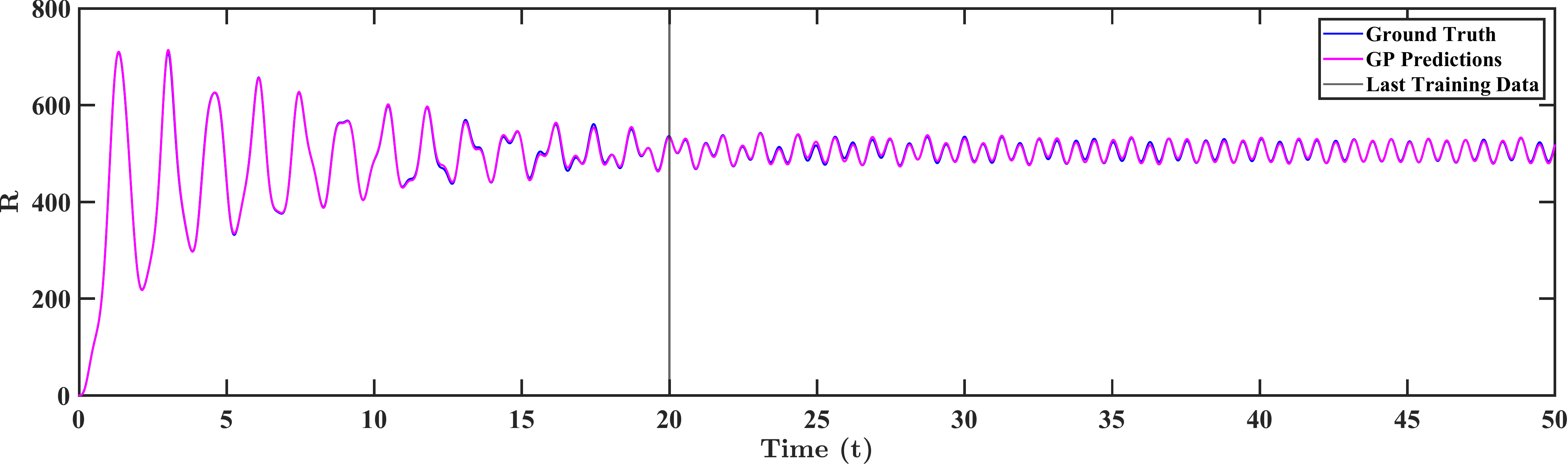}
	\caption{\centering Projected residual forces (magenta) compared against ground truth (blue) for Case-III. Training length for GPR is given up-to 20 seconds.}
	\label{dvp-gp}
\end{figure}

Fig. \ref{dvp-gp} shows the model-form error estimated using the proposed approach. Training data is provided for up-to 20 seconds and the results obtained matches almost exactly with the ground truth.

\begin{figure}[ht!]
	\centering
	\includegraphics[width = 1\textwidth]{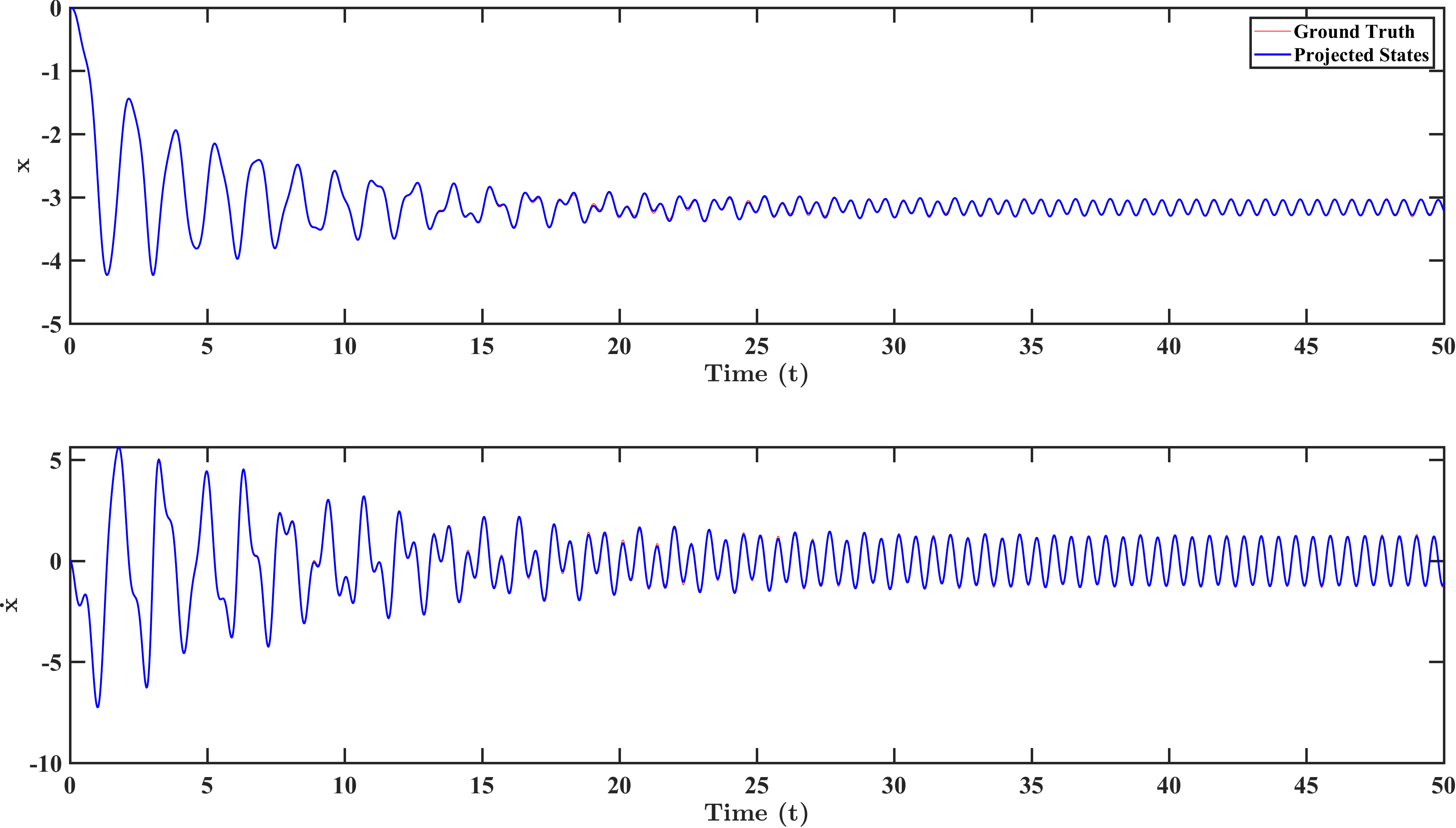}
	\caption{\centering Projected states (blue) compared against ground truth (red) when system is subjected to `same input' for case-III}
	\label{dvp-sf}
\end{figure}

\begin{figure}[ht!]
	\centering
	\includegraphics[width = 1\textwidth]{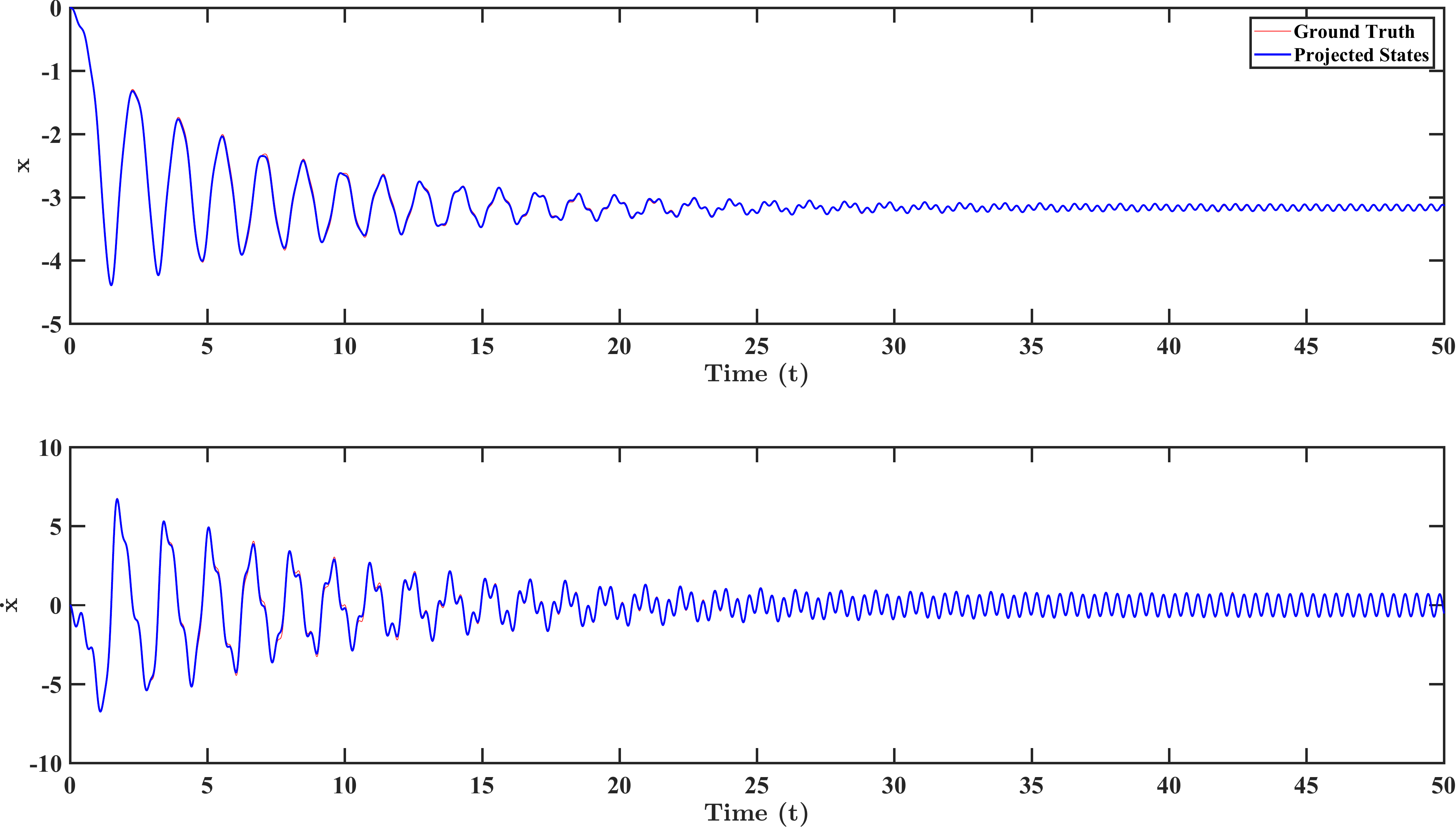}
	\caption{\centering Projected states (blue) compared against ground truth (red) when system is subjected to 'different input' for case-III}
	\label{dvp-df}
\end{figure}

Having showcased the performance of the proposed approach in identifying the model-form error, we proceed to examine its performance in predicting the state variables. To that end, we include the identified model-form error (in terms of GP) as an additional term into the known but approximate governing equation and solve the forward problem. Similar to previous examples, we consider two cases, one where the system is subjected to the same input as the training data and one where the system is subjected to different input. Fig. \ref{dvp-sf} and \ref{dvp-df} shows the results for projected states when system is subjected to same input and different input respectively. For both the cases, the projected states closely follow the ground truths, reflecting the efficacy of the proposed algorithm.


\section{Conclusion}\label{S5: C}
In this paper, we proposed a novel gray-box modeling approach for quantifying model-form uncertainty in nonlinear dynamical systems. The proposed approach blends known (but approximate) governing physical laws with data-driven machine learning algorithm. The primary idea is to treat the model-form error as a residual force and estimate it using input estimation approach. We propose using duel Bayesian filters for jointly estimating the input and the state vector. We argue that only identifying model-form error is not sufficient, and should be complemented with a framework allows merging the model-form error into the known but approximate governing equation so as to improve its predictive capability. To that end, we propose to express the identified model-form error as a function of the state-vector and use a machine learning algorithm to learn the mapping between the two. The trained machine learning model is then substituted into the known but approximate governing equation so as to improve its predictive capability. Although any machine learning algorithm can be used within the proposed approach, we have used Gaussian process regression \cite{nayek2019gaussian} in this study. It should be noted that while mapping estimated states to the residual force, either all the displacements and velocities can be used or only those can be selected which can effect the residual force in a meaningful way. 
For example if in a 5-DOF system, residual forces at third degree of freedom $R_3$ are to be mapped, displacements and velocities of only 2$^\text{nd}$, $3-$rd and $4-$th DOF may be required i.e. $R_3 = f(\bm {X_{2:4}},\bm{\dot X_{2:4}})$.
This approach can greatly reduce the computational requirements but may be employed only when a basic idea of original system is available.

In spite of the excellent results obtained for the examples presented, we note that the proposed framework can be further developed. For example, the two duel Bayesian filters used in this study are only conditionally stable hence the proposed framework will be able to produce results for a certain set of system parameters only. This condition is intensified by the fact that the non-linear systems being analyzed have an inherent tendency of becoming unstable when subjected of different inputs. Similarly, the proposed approach can also be applied to systems governed by partial differential equations. Applications of proposed algorithm in conjunction with different technologies can also be explored, where one possible use for the framework could be to merge it with digital twin technology, which in itself is an area of research with vast potential.

\section*{Acknowledgment}
SC acknowledges the financial support received from IIT Delhi in form of seed grant.
\appendix

\section{Bayesian filter model for linear dynamical systems}\label{app:1}
Discussed briefly here are the basics behind forming the filter model for the known MDOF system.
First case is for when the prop model for known system is selected as a linear dynamical system with governing equation as follows:
\begin{equation}
	\mathbf{M} \bm{\ddot X} + \widetilde{\mathbf{C}} \bm{\dot X} + \widetilde{\mathbf{K}} \bm X + \bm{R} = \bm F+\Sigma \bm W,
	\label{DE-linear-assumed model}
\end{equation}
Note that stochastic forces by nature are random and for the scope of the current study, their effect on dynamical system is considered as equivalent to process noise.
Thus there contribution in filter model is covered in process noise co-variance.
Dynamic model function for the filter model can then be obtained by following the procedure explained below.

The state vector for this case can be idealized as $\bm{\mathrm y} = [\bm X, \bm{\dot X}]^T$ and the unknown force vector will be simply $\bm R$.
Accelerations measured can be mathematically described as:
\begin{equation}
	\bm A = -\mathbf{M}^{-1}(\widetilde{\mathbf{K}}\bm X+\widetilde{\mathbf{C}}\bm{\dot X}+\bm R)
\end{equation}
Now, the governing equation \ref{DE-linear-assumed model} can then be rearranged as:
\begin{equation}
	\bm{\dot{\mathrm y}} = \mathbf{A_c}\bm{\mathrm y}+\mathbf{B_c}\bm F+\mathbf{C_c}\bm R,
\end{equation}
where
\begin{equation}
	{\mathbf{A_c}=\left[\begin{matrix} \bm 0 & \mathbf{I}\\ -\mathbf{M}^{-1}\widetilde{\mathbf{K}} & -\mathbf{M}^{-1}\widetilde{\mathbf{C}}\end{matrix}\right]}\text{ and } {\mathbf{B_c}=-\mathbf{C_c}=\left[\begin{matrix} \bm 0 \\ \mathbf{M}^{-1} \end{matrix}\right]}
	\label{E:LM:R}
\end{equation}
Discretization of Eq. (\ref{E:LM:R}) at a sampling period of $dt$ is as follows:
\begin{equation}
	\bm{\mathrm y}_k = (\mathbf{A_d}\bm{\mathrm y}+\mathbf{B_d}\bm F+\mathbf{C_d}\bm R)_{k-1},
\end{equation}
where $\mathbf{A_d} = exp(\mathbf{A_c}dt)$, $\mathbf{B_d}=[\mathbf{A_d}-\mathbf{I}]\mathbf{A_c}^{-1}\mathbf{B_c}$ and $\mathbf{C_d}=[\mathbf{A_d}-\mathbf{I}]\mathbf{A_c}^{-1}\mathbf{C_c}$. Similarly measurement model can be idealized as follows:
\begin{equation}
	\bm{\mathrm z}_k = (\mathbf{A_m}\bm{\mathrm y}+\mathbf{C_m}\bm R)_k
\end{equation}
where
\begin{equation}
	{\mathbf{A_m}=\left[\begin{matrix} -\mathbf{M}^{-1}\widetilde{\mathbf{K}} & -\mathbf{M}^{-1}\widetilde{\mathbf{C}}\end{matrix}\right]}\text{ and } {\mathbf{C_m}= -\mathbf{M}^{-1}}
\end{equation}
Hence final model for DKF\cite{dertimanis2019input} with additive noise can be written as:
\begin{equation}
	\begin{array}{c}
		\bm R_k = \bm R_{k-1}+\bm q^1_{k-1}\\
		\bm{\mathrm y}_k = (\mathbf{A_d}\bm{\mathrm y}+\mathbf{B_d}\bm F+\mathbf{C_d}\bm R)_{k-1}+\bm q^2_{k-1}\\
		\bm{\mathrm z}_k = (\mathbf{A_m}\bm{\mathrm y}+\mathbf{C_m}\bm R)_k+\bm r_k
	\end{array}
	\label{DKF:sample}
\end{equation}
where $\bm q^i$ and $\bm r$ are process noise and measurement noise as mentioned in Eq. (\ref{eq-dkf}).
Residual forces in Eq. (\ref{DKF:sample}) are modelled as random walk, giving the value of $\mathbf{T}$ as $[1]_{n\times1}$. The final model can be analyzed using Algorithm \ref{alg-2}.

\section{Bayesian filter model for nonlinear dynamical systems}\label{app:2}
When the known system is a non-linear system, DUKF is used.
The governing equation for such a system will be same as that mentioned in Eq. (\ref{E:KDS:GE}).
Effect of stochastic forces is again considered in the process noise.
The governing equation as a time derivative of $\mathrm{y}(=[\bm X, \bm{\dot X}]^T)$ can then be written as follows:
\begin{equation}
  \dot{\mathrm{y}} = \left[\begin{matrix}\bm{\dot X}\\\bm F-(\mathbf{\widetilde{C}}\bm{\dot X}+\mathbf{\widetilde{K}}\bm X+\bm{\widetilde{N}}+\bm R)\end{matrix}\right]
  \label{SS:NL known model}
\end{equation}
Eq. (\ref{SS:NL known model}) can be simply discretized as follows:
\begin{equation}
  \bm{\mathrm y}_k = (\bm{\mathrm y}+\bm a\,dt)_{k-1},
\end{equation}
where 
\begin{equation}
  \bm a = \left[\begin{matrix}\bm{\dot X}\\\bm F-\mathbf{M}^{-1}(\mathbf{\widetilde{C}}\bm{\dot X}+\mathbf{\widetilde{K}}\bm X+\bm{\widetilde{N}}+\bm R)\end{matrix}\right]
\end{equation}
A more precise discretization can be used if mandated by the problem under consideration. Accelerations for this case can numerically described as:
\begin{equation}
	\bm A = -\mathbf{M}^{-1}(\widetilde{\mathbf{K}}\bm X+\widetilde{\mathbf{C}}\bm{\dot X}+\widetilde{\bm N}+\bm R)
\end{equation}
If the forces are modelled as random walk, the dynamic and measurement model functions from Eq. (\ref{eq-dukf}) can be written as:
\begin{equation}
  \begin{matrix}
  f^1(\cdot) = \bm R_{k-1}\\
  f^2(\cdot) = (\bm{\mathrm y}+\bm a\,dt)_{k-1}\\
  h(\cdot) = -\mathbf{M}^{-1}(\widetilde{\mathbf{K}}\bm X+\widetilde{\mathbf{C}}\bm{\dot X}+\widetilde{\bm N}+\bm R)_k
  \end{matrix}
\end{equation}


\end{document}